\documentclass[runningheads]{llncs}
\usepackage{epsfig}
\usepackage{graphicx}
\usepackage[inkscapelatex=false]{svg}
\usepackage{amsfonts}
\usepackage{rotating}
\usepackage{booktabs}
\usepackage{adjustbox}
\usepackage{array}
\usepackage{placeins}
\usepackage{afterpage}
\usepackage{stfloats}
\usepackage{textcomp}
\usepackage{tikz}
\usepackage{comment}
\usepackage{amsmath,amssymb} 
\usepackage{color}
\usepackage{multirow}
\usepackage{subfigure}
\usepackage{mathrsfs}
\usepackage{graphicx}
\usepackage{float}

\DeclareUnicodeCharacter{2061}{}
\usepackage[pagebackref=true,breaklinks=true,letterpaper=true,colorlinks,bookmarks=false]{hyperref} 

\usepackage[width=122mm,left=12mm,paperwidth=146mm,height=193mm,top=12mm,paperheight=217mm]{geometry}

\newcommand{\figref}[1]{Figure~\ref{fig:#1}}
\newcommand{\tabref}[1]{Table~\ref{tab:#1}}

\newcolumntype{R}[2]{%
    >{\adjustbox{angle=#1,lap=\width-(#2)}\bgroup}%
    l%
    <{\egroup}%
}
\newcommand*\rot{\multicolumn{1}{R{45}{1em}}}

\newfloat{figtab}{htb}{fgtb}
\makeatletter
  \newcommand\figcaption{\def\@captype{figure}\caption}
  \newcommand\tabcaption{\def\@captype{table}\caption}
\makeatother
\begin{document}
\pagestyle{headings}
\mainmatter
\title{HFT: Lifting Perspective Representations via \\ Hybrid Feature Transformation} 

\titlerunning{Lifting Perspective Representations via Hybrid Feature Transformation}

\author{Jiayu Zou\inst{1} \quad Junrui Xiao\inst{1} \quad Zheng Zhu\inst{2} \quad Junjie Huang\inst{2} \quad Guan Huang\inst{2} \quad Dalong Du\inst{2} \quad Xingang Wang\inst{1}}

\institute{Institute of Automation, Chinese Academy of Sciences \and
PhiGent Robotics}
\authorrunning{Zou et al.}
\maketitle
\vspace{-8 mm}
\begin{abstract}
Autonomous driving requires accurate and detailed Bird's Eye View (BEV) semantic segmentation for decision making, which is one of the most challenging tasks for high-level scene perception. Feature transformation from frontal view to BEV is the pivotal technology for BEV semantic segmentation. Existing works can be roughly classified into two categories, \textit{i}.\textit{e}., Camera model-Based Feature Transformation (CBFT) and Camera model-Free Feature Transformation (CFFT). In this paper, we empirically analyze the vital differences between CBFT and CFFT. The former transforms features based on the flat-world assumption, which may cause distortion of regions lying above the ground plane. The latter is limited in the segmentation performance due to the absence of geometric priors and time-consuming computation. In order to reap the benefits and avoid the drawbacks of CBFT and CFFT, we propose a novel framework with a Hybrid Feature Transformation module (HFT). Specifically, we decouple the feature maps produced by HFT for estimating the layout of outdoor scenes in BEV. Furthermore, we design a mutual learning scheme to augment hybrid transformation by applying feature mimicking. Notably, extensive experiments demonstrate that with negligible extra overhead, HFT achieves a relative improvement of 13.3\% on the Argoverse dataset and 16.8\% on the KITTI 3D Object datasets compared to the best-performing existing method. The codes are available at \href{https://github.com/JiayuZou2020/HFT}{https://github.com/JiayuZou2020/HFT}.

\keywords{Autonomous Driving, Perspective Representations, Bird's Eye View}
\end{abstract}
\section{Introduction}
\quad In recent years, with the rapid development of autonomous driving technology, researchers make extensive investigations in 3D object detection \cite{chen2016monocular,3D1,3D2,3D3,3D4,chen2017multi}, vehicle behavior prediction \cite{BC0,BC1,BC2,BC3}, and scene perception \cite{dev1,dev2,dev5,dev4}. Autonomous vehicles demand a detailed, succinct and compact representation of their surrounding environments for path planning and obstacle avoidance \cite{dev9,dev11}. Bird's Eye View (BEV) representations satisfy the aforementioned requirements. Researchers usually re-represent the real world in the form of BEV instead of reconstructing the full 3D world in its entirety \cite{schulter2018learning,dev7,dev8}. BEV semantic maps become popular in capturing the scene layouts for both static and dynamic objects in autonomous driving scenarios. BEV semantic segmentation is the fundamental and vital basis for high-level motion prediction, lane detection \textit{et al}. Intuitively, reasonable BEV semantic segmentation requires the ability to holistically exploit images' spatial context and capture high-level semantic knowledge. Accurate scenes layout estimation for road, vehicle, pedestrian, and other objects provides meaningful information for making high-level decisions. Detailed and accurate BEV segmentation requires the ability of perceive holistically about the real world from sensor observations. \par  

To capture accurate BEV semantic maps, popular methods in industry community rely on expensive sensors such as LiDAR and radar \cite{beltran2018birdnet,li2016vehicle}. These methods require time-consuming computation to process cloud point data. Given the limited resolution and lack of sufficient semantic information of sensors, we focus on performing the BEV semantic segmentation by a single monocular RGB image \cite{chabot2017deep}. The existing methods of lifting the outdoor scenes from frontal view (FV) to BEV can be broadly divided into two categories, \textit{i}.\textit{e}., 1) Camera model-Based Feature Transformation (CBFT) \cite{reiher2020sim2real,roddick2020predicting,philion2020lift}, and 2) Camera model-Free Feature Transformation (CFFT) \cite{yang2021projecting,lu2019monocular,pan2020cross}. The former exploits the geometric priors to transform the coordinates from FV into BEV. CBFT uses Inverse Perspective Mapping (IPM) or its variants to capture geometric priors, and adopts deep Convolutional Neural Networks (CNN) to refine the coarse perspective features in top-down view. On the contrary, the CFFT simulates the global projection process without geometric prior, which can get rid of the limitation of the flat-world assumption. Due to their inherent drawbacks, both CBFT and CFFT methods fall into erroneous BEV maps and bring great challenges to accurate perception in BEV.\par

\begin{figure*}[htp]
    \setlength\tabcolsep{.01cm}
    \begin{minipage}{0.50\linewidth}
\begin{center}
    \begin{tabular}{ccccc}
        \includegraphics[height=1.3cm]{figures/essential_diff/6e7b/source_img.png}&
        \includegraphics[height=1.3cm]{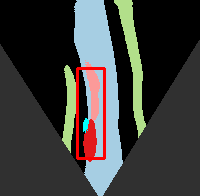}&
        \includegraphics[height=1.3cm]{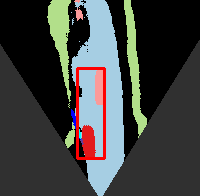}&
        \includegraphics[height=1.3cm]{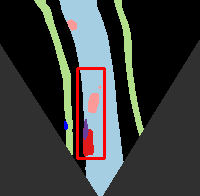}&
        \includegraphics[height=1.3cm]{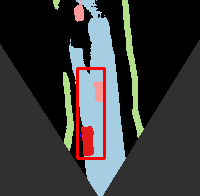}\\
        
        \includegraphics[height=1.3cm]{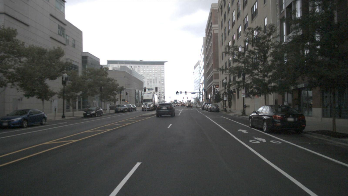}&
        \includegraphics[height=1.3cm]{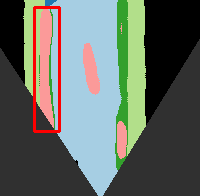}&
        \includegraphics[height=1.3cm]{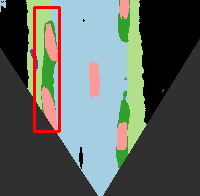}&
        \includegraphics[height=1.3cm]{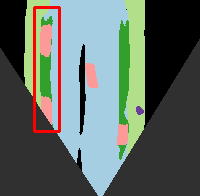}&
        \includegraphics[height=1.3cm]{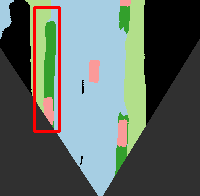}\\
    
        \includegraphics[height=1.3cm]{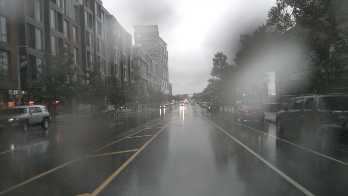}&
        \includegraphics[height=1.3cm]{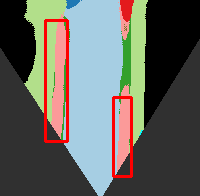}&
        \includegraphics[height=1.3cm]{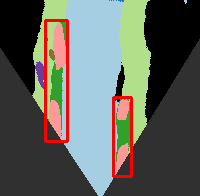}&
        \includegraphics[height=1.3cm]{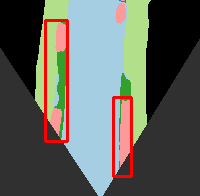}&
        \includegraphics[height=1.3cm]{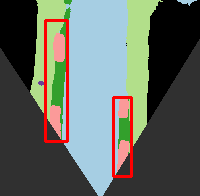}\\

        \tiny{FV Image} & \tiny{CFFT-20k}  & \tiny{CBFT-20k}  & \tiny{CFFT-200k} & \tiny{CBFT-200k} \\ 
    \end{tabular}
\end{center}
\end{minipage}
\hfill
\begin{minipage}{0.36\linewidth}
\begin{center}
    \includegraphics[width=1.0\linewidth,trim= 0 5 0 5,clip]{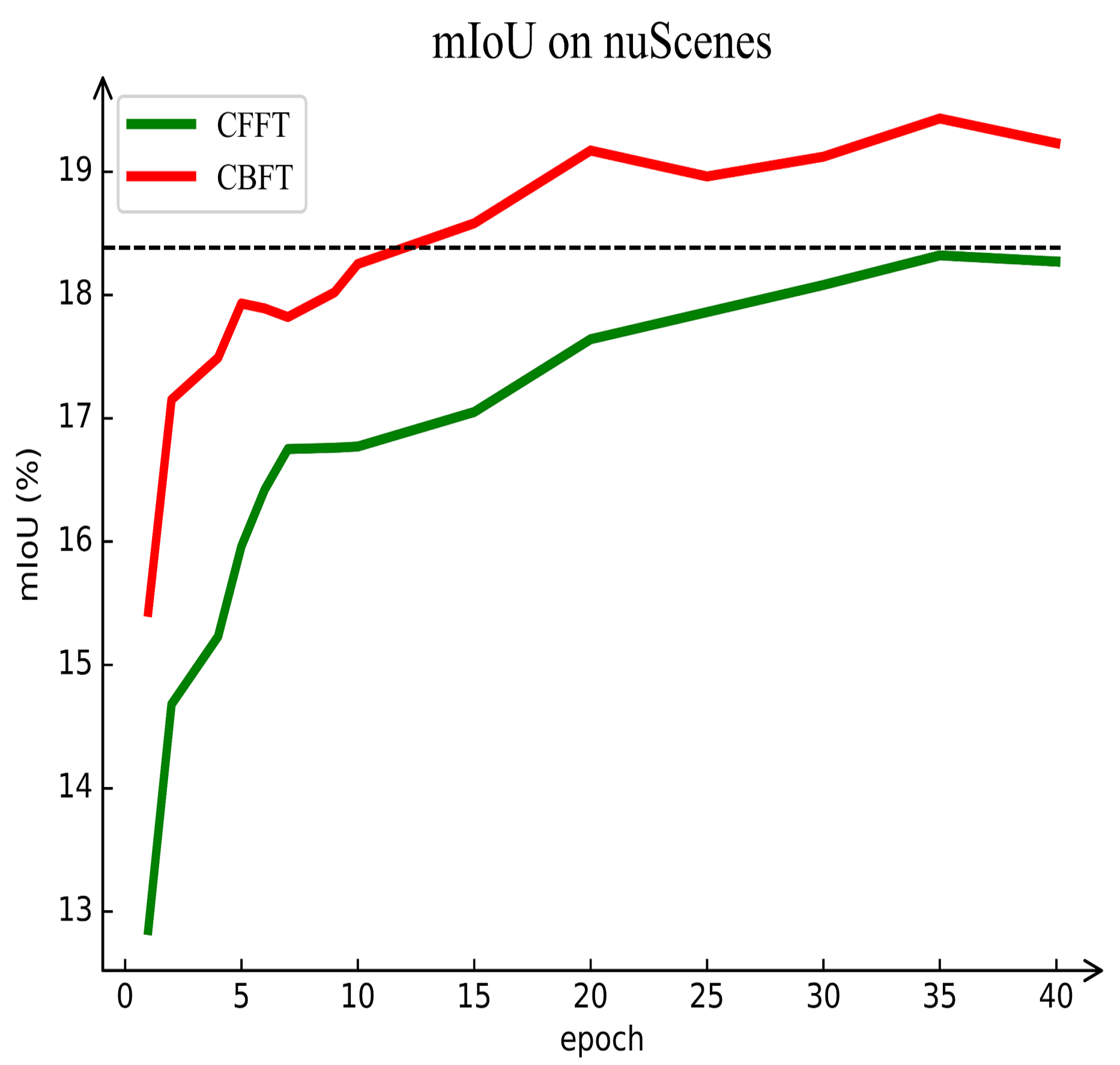}
\end{center}
\end{minipage}
    \caption{The differences between CBFT and CFFT. The images in the 2nd, 3rd 4th, and 5th column are semantic maps in BEV produced by PYVA \cite{yang2021projecting} trained for 20k iterations, PON \cite{roddick2020predicting} trained for 20k iterations, PYVA \cite{yang2021projecting} trained for 200k iterations, and PON \cite{roddick2020predicting} trained for 200k iterations, respectively.}
    \label{fig:essential diff}
\end{figure*}

To analyze the differences and problems of the aforementioned methods, we visualize the induced changes of the perspective transformation in the training phase on the nuScenes dataset. On the one hand, CBFT can handily handle the perspective feature transformation in flat regions such as local roads and carparks. CBFT strongly relies on the flat-world assumption, which is the basis of IPM or its variants. As a consequence, there is a distortion in CBFT for those regions above the ground. Provided camera parameters, geometric priors may help CBFT converge faster than CFFT, as revealed in \figref{essential diff}. On the other hand, CFFT ignores the geometric priors and employs MLP or attention mechanism to capture global spatial correlation. To this end, CFFT may achieve better performance in areas lying above the ground. The above findings suggest that the geometric priors help the model converge, while the global spatial correlation serves to get rid of the limitations of the flat-world assumption.\par

Based on the aforementioned analysis, we propose the Hybrid Feature Transformation (HFT) module, which consists of two different branches: Geometric Transformer and Global Transformer, as shown in \figref{overall}. Specifically, the Geometric Transformer utilizes an IPM-style layer to obtain rough BEV feature maps in the early training stage. Provided camera parameters, Geometric Transformer benefits from geometric priors. Global Transformer is designed to transform FV to top-down view with multi-layer perceptron (MLP) \cite{multilayer} or attention mechanism \cite{attentionisall}, which can globally model the correlation between different views. Global Transformer works well in capturing the global context over the images. We present more detailed analysis in section 3.1. To fully exploit the increased modeling capacity of HFT, we introduce mutual learning \cite{hinton2015distilling,dual,coo} to push two different branches to learn potential representation from each other.\par
From experimental results, we show that the proposed HFT can significantly facilitate the accuracy of BEV semantic segmentation. Without bells and whistles, HFT achieves a relative improvement of 13.3\% and 16.8\% than previous state-of-the-art methods on the Argoverse and KITTI 3D Object datasets respectively. With negligible extra overhead than CBFT and even less computing budget than CFFT, HFT obtains considerable improvement in segmentation performance. Compared with View Parsing Network (VPN) \cite{pan2020cross}, HFT reduces the number of parameters by 31.2\% from 50.74M to 34.92M and still achieves a large relative 21.1\% improvement of mIoU. In ablation study, we illustrate the effectiveness of HFT framework and its key components.\par
The contributions of this paper can be summarized as follows:

\begin{itemize}
\item[$\bullet$] We empirically discuss the vital differences between CBFT and CFFT. To the best of our knowledge, we are the first to point out that both the geometric priors in CBFT and the global spatial correlation in CFFT are important for BEV semantic segmentation.
\item[$\bullet$] We propose a novel end-to-end learning framework, named HFT, to reap the benefits and avoid the drawbacks of CBFT and CFFT to construct an accurate BEV semantic map using only a monocular FV image. 
\item[$\bullet$] HFT is evaluated on public benchmarks and achieves state-of-the-art performance with negligible computing budget, \textit{i}.\textit{e}., at least relative 13.3\% and 16.8\% improvement than previous methods on the Argoverse and KITTI 3D Object datasets respectively.
\end{itemize}

\section{Related Works}
\quad In autonomous driving tasks, BEV representation occupies a crucial position \cite{hu2021fiery,li2021hdmapnet,du2018general,wirges2018object,qi2018frustum,poirson2016fast}. Several works have been proposed to obtain accurate BEV semantic maps from a single 2D FV and these works can be roughly classified into CBFT and CFFT. \par

\subsection{Camera model-Based Feature Transformation}
\quad CBFT methods account for the geometry of the scene to transform the original FV image into the BEV space. Camera to Birds' Eye View (CAM2BEV) \cite{reiher2020sim2real} employs IPM for perspective transformation, but the applicable scenarios of IPM are limited to pixels above the ground plane. Pyramid Occupancy Networks (PON) \cite{roddick2020predicting} encodes the image onto the BEV space with pre-defined depth, but it induces large deviations and limits segmentation performance on the classes that spread over the entire image, \textit{e}.\textit{g}., curbs and lanes. Lift-Splat-Shoot (LSS) \cite{philion2020lift} estimates the probability distribution of the depth and projects FV features back into a voxel grid for perspective transformation. However, the variation of depth across different categories causes degradation of segmentation performance for many classes. \par
\subsection{Camera model-Free Feature Transformation}  
\quad Contrary to the CBFT methods, CFFT fully relies on the model's representation ability to lift perspective representations. Projecting Your View Attentively (PYVA) \cite{yang2021projecting} simply ignores geometric priors and employs the attention mechanism to learn the warping from the FV to BEV. However, PYVA \cite{yang2021projecting} fails to estimate the geometry of layouts and the number of objects accurately, making the network converge slowly. VED \cite{lu2019monocular} utilizes a variational autoencoder (VAE) architecture to encode the FV features and decode them into BEV space directly. The properties of VAE cause the sharp edges, being not conducive to obtaining semantics with clear boundaries. VPN \cite{pan2020cross} adopts MLP to transform FV features into BEV space. However, discarding geometric priors forces the network to approximate the transformation and makes the output coarser. Our proposed HFT framework alleviates the above problems and enables the model to obtain more accurate warping.

\section{Method}

\begin{figure*}[]
    \centering
    \includegraphics[width=\linewidth,trim= 0 10 0 5,clip]{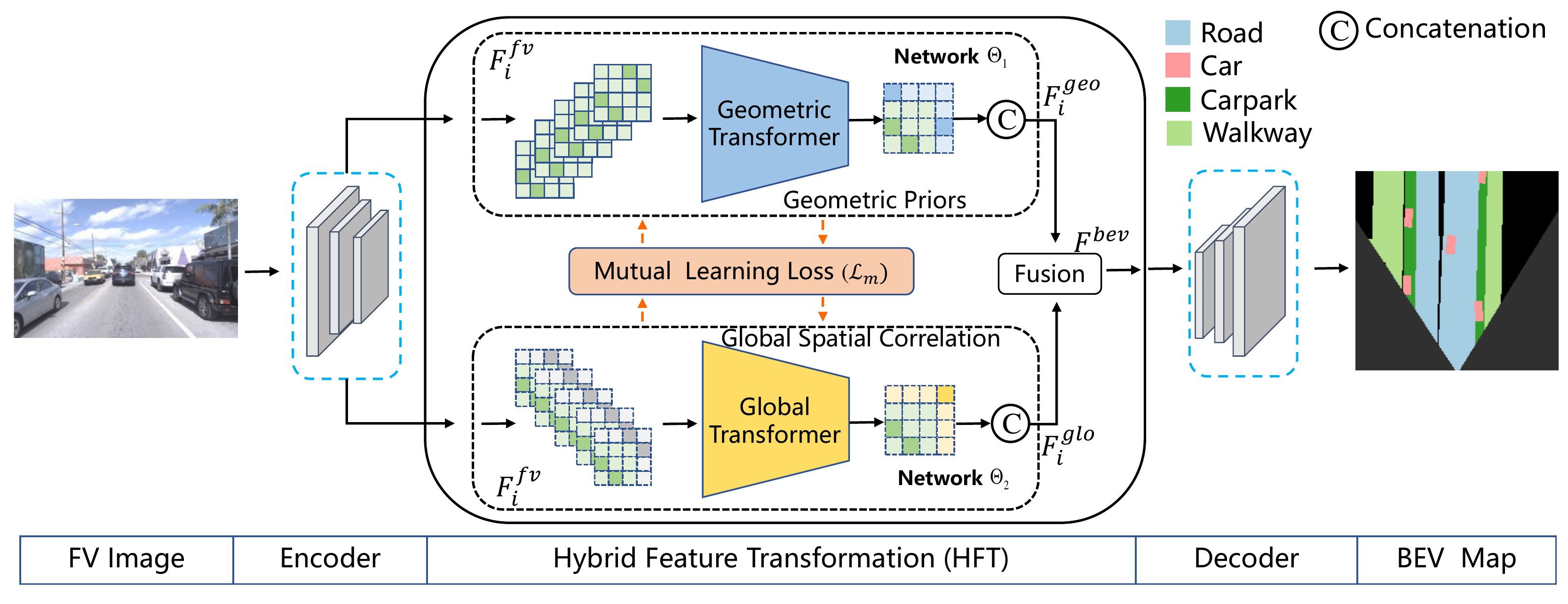}
    \scriptsize
    \caption{Architecture diagram showing an overview of our proposed HFT framework. (1) FV image encoder extracts high-level features at multiple resolutions. (2) Geometric Transformer branch employs camera parameters to attain perspective feature transformation. (3) Global Transformer branch learns perspective transformation without relying on camera parameters. (4) A mutual learning scheme is conducted to push two branches to learn better perspective representation. (5) BEV Image Decoder predicts the final semantic occupancy probabilities in every grid.}
    \label{fig:overall}
\end{figure*}

\quad In this section, based on the analysis of the differences between CBFT and CFFT, we present HFT architecture for BEV semantic segmentation. As illustrated in \figref{overall}, we take an image $I_i\in \mathbb{R}^{H\times W\times3}$ with an intrinsic matrix $M_i\in \mathbb{R}^{3\times3}$ as inputs. We strive for a BEV semantic representation $\mathcal{F}_i^{bev}\in \mathbb{R}^{C\times Z\times W}$, where $C$ is the number of categories and $(Z,W)$ is the reference coordinates. It's noteworthy that we use only a monocular FV image without the reliance on cloud point data.\par

\subsection{Difference Analysis}
\quad To explore the differences between CBFT and CFFT, we study the evolution of mean Intersection over Union (mIoU) scores on the nuScenes dataset. We select PON \cite{roddick2020predicting} and PYVA \cite{yang2021projecting}, as the representative for CBFT and CFFT, repectively. As illustrated in \figref{essential diff}, PON \cite{roddick2020predicting} yields defective segmentation results in regions that are above the ground, \textit{e}.\textit{g}, ramps and curbs. What's more, the evolution of segmentation performance manifests that PYVA \cite{yang2021projecting} converges slower than PON \cite{roddick2020predicting}.\par
The differences between CBFT and CFFT are summarized as follows. CBFT exploits the geometric priors explicitly by incorporating the camera intrinsics into the view transformer module. Such approaches follow the IPM to project image features from FV to BEV. Reasonable semantic segmentation results from the camera coordinate system to the pixel coordinate system can be obtained on the premise that the optical axis of the camera is parallel to the ground. However, as for objects lying above the ground plane, CBFT is prone to result in inappropriate semantic maps. In other words, the flat-world assumption in IPM hinders segmentation performance in areas lying above the ground. We take \figref{essential diff} for illustration, CBFT (PON \cite{roddick2020predicting}) for CBFT) generates an oval BEV semantic map for a car, whose semantics should be rectangular.\par
The main component of CFFT is MLP or attention mechanism, which works well in capturing global correlation in a pixel-wise manner. Thus, CFFT benefits from global spatial correlation. However, CFFT fails to yield BEV semantic maps with clear shapes or edges. The mIoU curve demonstrates that CFFT converges slower than CBFT. The key factor gives raise to the performance gap between CBFT and CFFT is whether the feature transformation employs the geometric priors. The visualization result of CFFT reveals that lacking geometric priors leads to imprecisely estimating the layout geometry.\par 
The above analysis suggests that both the flat-world assumption in CBFT, and lack of geometric priors in CFFT limit the accuracy and effectiveness in overall BEV semantic segmentation. The vital differences between CBFT and CFFT inspire us to design a novel framework.\par

\subsection{Network Overview}
\quad We follow the principle of reaping the benefits of both CBFT and CFFT while avoiding their drawbacks. In proposed HFT network, three kinds of modules are conducted in succession: a shared image-view encoder, HFT module and a semantic decode head. The shared image-view encoder adopts a shared backbone to extract FPN-style features with multiple scales. HFT module takes the high-level FV features as input and independently conducts two feature transformation. Furthermore, HFT employs a mutual learning strategy to pushing both CBFT branch (Geometric Transformer) and CFFT branch (Global Transformer) to learn more appropriate representations from each other. Thereafter, the semantic decode head deals with the semantic segmentation task for both dynamic and static elements in BEV.
\subsection{Hybrid Feature Transformation}
\quad Given the coordinate system in FV space is not the same as that in BEV space, there is a large amount of unaligned image content during perspective transformation. We design a novel transformation module, named HFT. Compared with CBFT and CFFT, HFT takes both geometric priors and global spatial correlation into account by feature mimicking, offering a new perspective towards perspective representations. HFT with a modular design consists of three kinds of modules: Global Transformer, Geometric Transformer, and mutual learning scheme. Both Global Transformer branch and Geometric Transformer branch independently estimate semantic probability occupancy in BEV space. Mutual learning measures the similarity between two branches and strives for more precise lifting perspective representations by feature mimicking. Finally, we conduct regularization on the semantic probability occupancy maps under different depth extents and concatenate feature maps $\mathcal{F}_i^{bev}$ to form the final BEV features $\mathcal{F}^{bev}$.\par 
\textbf{Geometric Transformer.} Geometric Transformer employs camera intrinsic matrices $M_i$ to produce perspective projection features $\mathcal{F}_i^{ipm}$ with rich geometric priors. Specifically, we flatten the FV feature map $\mathcal{F}_i^{fv}$ in height dimension. Following PON \cite{roddick2020predicting}, we adopt the IPM algorithm and use a resampling layer $C(x)$ to warp it into predefined depth extents. Concatenating feature maps in different depth ranges forms the final BEV features $\mathcal{F}^{geo}$. We denote the flatten and concatenation operation as $Flat$ and $Concat$, respectively.\par
\begin{equation}
\mathcal{F}^{geo} = Concat(\mathcal{F}_i^{geo}) = Concat(C(Flat(\mathcal{F}_i^{fv}),M_i))
\end{equation}
\quad \textbf{Global Transformer.} Different from Geometric Transformer, Global Transformer ignores geometric priors and thus gets rid of the flat-world assumption. The key component of Global Transformer is MLP or attention mechanism, which has a strong ability of capture global context. This property may lead to better performance on those classes above on the ground. To this end, we warp the 2D features $\mathcal{F}_i^{fv}$ from backbone into a volumetric grid $\mathcal{F}_i^{bev}$. Subsequently, we generate the spatial occupancy relation $R(x)$ \cite{yang2021projecting} to get feature maps $\mathcal{F}^{glo}$ in BEV. We denote the warping function from FV to BEV as $F$.\par
\begin{equation}
\mathcal{F}^{glo} = Concat(\mathcal{F}_i^{glo}) = Concat(R(F(\mathcal{F}_i^{fv})))
\end{equation}
\textbf{Mutual Learning.} For reaping the benefits of both the rich geometric priors of Geometric Transformer and the strong global spatial correlation of Global Transformer, we introduce mutual learning and feature mimicking into BEV semantic segmentation. Inspired by Deep Mutual Learning (DML) \cite{Zhang2018DeepML,lan2018knowledge,kim2021feature,phuong2019distillation} which uses feature maps to mutually distill knowledge from two models, we design a scheme to push the performance boundary of both branches by designing $\mathcal{L}_{m}$ which is described in section 3.4.

\subsection{Loss Function}
\quad The loss function consists of three parts: $\mathcal{L}_{s}$, $\mathcal{L}_{u}$, and $\mathcal{L}_{m}$. The first loss $\mathcal{L}_{s}$ is the cross-entropy loss between final semantic representation and pixel-wise annotations with hard mining. The second loss $\mathcal{L}_{u}$ is an uncertainty loss \cite{roddick2020predicting} for predicting the uncertainty of ambiguous areas in two corresponding BEV features. $\mathcal{L}_{m}$ indicates the mutual learning loss between two transformation branches in a pixel-wise manner. These three aspects are described as follows. \par
 \textbf{Semantic Loss.} Inspired by focal loss \cite{focal}, we modify the cross-entropy loss to address the class imbalance problem by increasing the weights of the infrequently occurring classes such as pedestrian and motorcycle. For each prediction with a classification score $c_i$, it has a corresponding binary label $y_i$ and semantic loss is written as
\begin{equation}
\mathcal{L}_{s} =\frac{1}{N_{pos}}[-\sum_i^{N_{pos}}w_i y_i\log c_i-\sum_i^{N_{neg}}(1-w_i)(1-y_i)\log⁡(1-c_i)]
\end{equation}
where $w_i$ is the weight of a class, which is computed as the inverse square root of its relative frequency.\par
 \textbf{Uncertainty Loss.}  As discussed in HFT module, Geometric Transformer and Global Transformer with their respective feature maps may conflict in BEV semantic maps. We adopt uncertainty loss to tackle the ambiguous regions by predicting a high uncertainty score, which is represented as
 \begin{equation}
\mathcal{L}_{u} =1-c_i\log c_i
\end{equation}
\quad \textbf{Mutual Learning Loss.} We design the mutual learning loss to measure the similarity of Geometric Transformer and Global Transformer, which can facilitate the flow of different lifted features. The mutual learning loss is formulated as
\begin{equation}
\mathcal{L}_{m} =\lambda_1||\mathcal{F}^{geo}-\mathcal{F}^{glo}||_2+\sum_i^{N_{fea}}\lambda_2||\mathcal{F}_i^{geo}-\mathcal{F}_i^{glo}||_2
\end{equation}
where $N_{fea}$ means the number of BEV semantic maps with respect to different depth extents. In practice, $\lambda_1$ and $\lambda_2$ are set as 0.05 and 0.01.\par
Combining the aforementioned three parts, the total loss function is represented as\begin{equation}
\mathcal{L}_{tot}=\mathcal{L}_{s}+\alpha \mathcal{L}_{u}+ \beta \mathcal{L}_{m}
\end{equation}where $\alpha$ = 0.001 and $\beta$ = 1.0 are weighting factors to balance the weights of the uncertainty loss and the mutual learning loss, respectively.\par

\section{Experiment}
\quad In this section, we conduct extensive experiments over challenging scenarios and compare HFT against the previous state-of-the-art methods on public benchmarks. Further ablation experiments are conducted to delve into the proposed framework.
\subsection{Datasets}
\quad In order to verify the effectiveness of our approach, we evaluate the performance in three aspects, \textit{i}.\textit{e}. static, dynamic, and hybrid scene layout estimation. We evaluate HFT framework on five different datasets, \textit{i}.\textit{e}., KITTI Raw, KITTI Odometry, KITTI 3D Object \cite{geiger2012we,behley2019semantickitti}, nuScenes \cite{caesar2020nuscenes}, and Argoverse \cite{chang2019argoverse}. The task of static scene layout estimation is performed on the KITTI Raw and KITTI Odometry. Extensive experiments on the KITTI 3D Object benchmark reveal semantic segmentation performance for dynamic vehicles. Furthermore, we conduct experiments for hybrid scene layout estimation on the Argoverse and nuScenes, which provide complex scenarios of both staic and dynamic elements.\par

To verify the feasibility on static or dynamic classes, we evaluate the segmentation performance on KITTI datasets. The KITTI Raw \cite{geiger2012we} generates ground truth segmentation by registering the depth and semantic segmentation of LiDAR scans. The KITTI Odometry \cite{behley2019semantickitti} provides 22442 well-annotated FV images and BEV masks. For comparison with existing 3D vehicle segmentation approaches \cite{lu2019monocular,chen2016monocular,roddick2020predicting,yang2021projecting,3D3,pan2020cross}, we evaluate performance on the KITTI 3D Object \cite{geiger2012we} benchmark.\par
The nuScenes dataset includes 1000 autonomous driving scenes, whose source data is collected with six surround-view cameras, five radar sensors and a LiDAR sensor. We only choose the images captured by the FV camera as our framework’s input.\par
In addition, we compare methods on Argoverse dataset, which includes images and point cloud data with seven surround view cameras, two stereo cameras and two LiDAR sensors. In practice, we evaluate performance on Argoverse split used in  \cite{schulter2018learning}, \textit{i}.\textit{e}., 6723 training images and 2418 validation images.\par
We employ mIoU and mean Average Precision (mAP) as the evaluation metrics. For hybrid scene estimation, we design a novel evaluation metric $BamIoU$ (Balanced mean Intersection over Union). $BamIoU$ consists of the normalized weighted sum of $IoU$ for $N_{st}$ categories of static elements and $N_{dy}$ categories of dynamic elements, which is defined as follows. $BamIoU$ is designed to measure the comprehensive segmentation performance for dynamic and static elements (higher is better).\par
\begin{equation}
BamIoU = \sum_i^{N_{st}}w_iIoU_{st_i} +\sum_j^{N_{dy}}w_jIoU_{dy_j}
\end{equation}
where $\sum_i^{N_{st}}w_i=1,\sum_j^{N_{dy}}w_j=1$.
\subsection{Implementation Details}
\quad \textbf{Training Parameters.} Models are trained by AdamW \cite{loshchilov2018fixing} optimizer, in which gradient clip is exploited with learning rate 2e-4. The total batch size is 48 on 4 NVIDIA GeForce RTX 3090 GPUs. For shared backbone, we initialize the SwinTransformer \cite{liu2021swin} backbone with weights pre-trained on the ImageNet \cite{Russakovsky2015ImageNetLS} and apply a step learning rate policy which drops the learning rate at 17 and 20 epochs by a factor of 0.1. The total schedule is terminated within 40 epochs for KITTI Raw, KITTI Odometry, KITTI 3D Object, and Argoverse datasets. Models terminate training within 55 epochs on the nuScenes dataset. We use the PyTorch \cite{Paszke2019PyTorchAI} and MMSegmentation \cite{mmseg2020} framework.\par
 \textbf{Preprocessing.} By default in the training process, the input images are resized to 1024$\times$1024 resolution. Training images are augmented by a probability of 50\% of random horizontal flipping, random perturbations of the image brightness, contrast, and saturation.\par
\begin{table*}[b]
\centering
\caption{Intersection over Union scores (\%) on the KITTI datasets.}
\label{tab:Kitti raw and odometry}
\scriptsize
\begin{tabular}{@{}c|cc|cc|cc@{}}
\toprule
KITTI & \multicolumn{2}{|c|}{Raw} & \multicolumn{2}{|c|}{Odometry}& \multicolumn{2}{|c}{3D Object}\\
\midrule
Method & mIoU & mAP & mIoU & mAP & mIoU & mAP\\
\midrule
MonoOccupancy \cite{lu2019monocular} & 58.41 & 66.01 & 65.74 & 67.84&20.45&22.29 \\
Mono3D \cite{chen2016monocular} & 59.58 & 79.07 & 66.81 & 81.79 &17.11&26.62\\
PYVA \cite{yang2021projecting} & 65.70 & \textbf{81.62} & 78.19 & 85.55&29.52&36.86  \\
PON \cite{roddick2020predicting} & 60.47 &  77.45 &  70.92 &  76.27&26.78&44.50 \\
OFT \cite{3D3}&-&-&-&-&25.34&34.69\\
VPN \cite{pan2020cross}&-&-&-&-&26.52&35.54\\
\midrule
Ours & \textbf{66.29} & 80.20 & \textbf{79.42} & \textbf{87.74}& \textbf{34.49} & \textbf{46.89} \\
\bottomrule
\end{tabular}
\end{table*}
\subsection{Benchmark Results}

\quad \textbf{Static Scene Layout Estimation.} To evaluate the performance in the task of static scene layout estimation, we compare HFT against Monocular semantic Occupancy (MonoOcc) \cite{lu2019monocular}, Monocular 3D (Mono3D) \cite{chen2016monocular}, PYVA \cite{yang2021projecting} and PON \cite{roddick2020predicting}. \tabref{Kitti raw and odometry} summarizes the performance of existing approaches on the KITTI Raw and KITTI Odometry benchmarks. We densify the sparse semantic labels and compare all methods under the same training protocol. As observed, HFT model ranks first among all the existing baselines by large margins on both datasets. HFT achieves the highest mIoU of 66.29\% and a competitive mAP of 80.20\% than concurrent models in KITTI Raw benchmark. Moreover, we observe a substantial improvement on the KITTI Odometry dataset in the mIoU and mAP when compared to both CBFT and CFFT baselines.\par

\textbf{Dynamic Scene Layout Estimation.} Taking the variability of scales and mobility into account, scene layout estimation for dynamic elements in BEV is a more challenging task than static ones. Note that, Mono3D \cite{chen2016monocular} is a two-stage method and OFT \cite{3D3} employs a parameter-heavy transformer which slows the speed down significantly. As shown in \tabref{Kitti raw and odometry}, HFT peaks both mIoU and mAP by a large margin, \textit{i}.\textit{e}, a mIoU score of 34.49\% and a mAP score of 46.89\%.\par
\begin{table*}[t!]
\centering
\caption{Intersection over Union scores (\%) on the Argoverse dataset.}
\label{tab:argoverse}
\scriptsize
\begin{tabular}{@{}c|cccccccc|c@{}}
\toprule
Method & Drivable & Vehicle & Pedest. & Large veh. & Bicycle & Bus & Trailer & Motorcy. &  mIoU\\
\midrule
IPM & 43.7 & 7.5 & 1.5 & - & 0.4 & 7.4 & - & 0.8 & -  \\
Unproj. & 33.0 & 12.7 & 3.3 & - & 1.1 & 20.6 & - & 1.6 & -  \\
VED \cite{lu2019monocular} & 62.9 & 14.0 & 1.0 & 3.9 & 0.0 & 12.3 & 1.3 & 0.0 & 11.9  \\
VPN \cite{pan2020cross} & 64.9 & 23.9 & 6.2 & 9.7 & 0.9 & 3.0 & 0.4 & 1.9 & 13.9  \\
PYVA \cite{yang2021projecting} & 78.95 & 33.91 & 6.87 & 18.29 & 6.1 & 32.5 & 32.39 & 1.01 & 26.25 \\
PON \cite{roddick2020predicting} & 65.70 & 27.72 & 6.56 & 8.08 & 0.25 & 19.87 & 16.49 & 0.16 & 18.10  \\
\midrule
Ours & \textbf{78.98} & \textbf{38.79} & \textbf{9.00} & \textbf{23.09} & \textbf{7.69} & \textbf{34.67} & \textbf{38.15} & \textbf{7.48} & \textbf{29.73} \\
\bottomrule
\end{tabular}
\end{table*}
\begin{table*}[b]
\centering
\caption{Balanced mean Intersection over Union scores (\%) on the nuScenes and Argoverse datasets.}
\label{tab:nuscenes and argoverse}
\scriptsize
\begin{tabular}{@{}c|ccc|ccc @{}}
\toprule
Dataset & \multicolumn{3}{|c|}{nuScenes} & \multicolumn{3}{|c}{Argoverse}\\
\midrule
Method &$ IoU_{st}$ & $IoU_{dy}$ & $BamIoU$ & $IoU_{st}$ &$ IoU_{dy}$ & $BamIoU$\\
\midrule
VED \cite{lu2019monocular}  & 0.79 & 0.10 &  0.89  & 0.63 & 0.36 & 0.99 \\
VPN \cite{pan2020cross} & - & - &  - & 0.65 & 0.64 & 1.29\\
PYVA \cite{yang2021projecting} & 1.42 & 0.91 &  2.33 & \textbf{0.79} & 1.68 & 2.47\\
PON \cite{roddick2020predicting}& 1.92 & 1.41 &  3.33 & 0.66 & 0.96 & 1.62 \\
\midrule
Ours & \textbf{2.06} & \textbf{1.67} &  \textbf{3.72} & \textbf{0.79} &\textbf{2.23} & \textbf{3.02}\\
\bottomrule
\end{tabular}
\end{table*}
\textbf{Hybrid Scene Layout Estimation.} The hybrid scene layout estimation is an essential sub-task for autonomous driving, which contains estimation of both static and dynamic elements. For a fair comparison with concurrent methods, we apply the train-validation splits and ground-truth generation protocol provided by \cite{chang2019argoverse,roddick2020predicting}. As shown in \tabref{argoverse}, HFT achieves the highest IoU score in all categories of objects on Argoverse dataset,\textit{i}.\textit{e}, 29.73\% mIoU score, which has more than a relative 13.3\% improvement to previous methods. Especially for dynamic classes, HFT shows strong ability of semantic representation in BEV space and significantly facilitates performance on these kinds of objects. Furthermore, as listed in \tabref{nuscenes}, HFT scores 21.3\% over 14 classes on nuScenes dataset, which offers more than a relative 10.4\% improvement on the baselines. HFT achieves state-of-the-art IoU scores in most classes and competitive result in drivable area. It's worth noting that HFT outperforms other existing models in infrequently occurring classes by a large margin, \textit{e}.\textit{g}., pedestrian and bicycle. As listed in \tabref{nuscenes and argoverse}, HFT model ranks first in static and dynamic elements on both nuScenes and Argoverse benchmarks.\\

\begin{table*}[t]
\centering
\caption{Intersection over Union scores (\%) on the nuScenes dataset.}
\label{tab:nuscenes}
\scriptsize
\begin{tabular}{@{}c|cccccccccccccc|c@{}}
\toprule
Method & \rot{Drivable} & \rot{Ped. crossing} & \rot{Walkway} & \rot{Carpark} & \rot{Car} & \rot{Truck} & \rot{Bus} & \rot{Trailer} & \rot{Constr. veh.} & \rot{Pedestrian} & \rot{Motorcycle} & \rot{Bicycle} & \rot{Traf. Cone} & \rot{Barrier} & \rot{Mean} \\
\midrule
IPM & 40.1 & - & 14.0 & - & 4.9 & - & 3.0 & - & - &  0.6 & 0.8 & 0.2 & - & - & - \\
Unproj. & 27.1 & - & 14.1 & - & 11.3 & - & 6.7 & - & - & 2.2 & 2.8 & 1.3 & - & - & -  \\
VED \cite{lu2019monocular}  & 54.7 & 12.0 & 20.7 & 13.5 & 8.8 & 0.2 & 0.0 & 7.4 & 0.0 & 0.0 & 0.0 & 0.0 & 0.0 & 4.0 & 8.7 \\
PYVA \cite{yang2021projecting} & \textbf{56.2} & 26.4 & 32.2 & 21.3 & 19.3 & 13.2 & 21.4 & 12.5 & \textbf{7.4} & 4.2 & 3.5 & 4.3 & 2.0 & 6.3 & 16.4 \\ 
PON \cite{roddick2020predicting} & 54.6 & 32.0 & 33.7 &  19.4 & 30.2 & 16.0 & 23.5 & 16.0 & 3.5 & 6.8 & 9.1 & 9.8 & 4.9 & 10.9 & 19.3 \\ \midrule
Ours & 55.9 & \textbf{35.6} & \textbf{35.4} & \textbf{23.2} & \textbf{30.6} & \textbf{19.1} & \textbf{27.3} & \textbf{18.6} & 5.8 & \textbf{7.9} & \textbf{11.1} & \textbf{11.7} & \textbf{5.1} & \textbf{11.4} & \textbf{21.3} \\ 
\bottomrule
\end{tabular}
\end{table*}
\subsection{Ablation Studies}
\textbf{The Effectiveness of HFT Framework.} To explore the effectiveness of HFT framework in perspective representations, we study the performance under different configurations on KITTI 3D Object dataset. As illustrated in \tabref{mutuallearning_table}, we compare the semantic segmentation performance of CBFT, CFFT and HFT methods with different components. We demonstrate the effectiveness of HFT in three folds.\par
\begin{figtab}
\setlength\tabcolsep{.01cm}
 \begin{minipage}[b]{0.56\linewidth}
    \centering
    \scriptsize
\begin{tabular}{@{}c|c|c|c|c|c @{}}
\toprule
Model & MLS & CBFT & CFFT & \#Params & mIoU (\%) \\
\midrule
PYVA \cite{yang2021projecting} & - & - & \checkmark & 47.12M & 29.27\\
VPN \cite{pan2020cross} & - & - & \checkmark & 50.74M & 28.34\\
PON \cite{roddick2020predicting} & - & \checkmark & - & 37.42M & 26.78\\
LSS \cite{philion2020lift} & - & \checkmark & - & 29.55M & 30.98\\
\midrule
\multirow{2}{*}{HFT-A} & - & \multirow{2}{*}{PON \cite{roddick2020predicting}} & \multirow{2}{*}{PYVA \cite{yang2021projecting}} & \multirow{2}{*}{40.96M} & 31.65\\ \cline{2-2} \cline{6-6} & \checkmark & & & & 34.49\\\midrule
\multirow{2}{*}{HFT-B} & - & \multirow{2}{*}{PON \cite{roddick2020predicting}} & \multirow{2}{*}{VPN \cite{pan2020cross}}  & \multirow{2}{*}{43.97M} & 33.52\\ \cline{2-2} \cline{6-6} & \checkmark & & & & 35.34\\ \midrule
\multirow{2}{*}{HFT-C} & -   & \multirow{2}{*}{LSS \cite{philion2020lift}} & \multirow{2}{*}{PYVA \cite{yang2021projecting}} & \multirow{2}{*}{37.45M} & 32.46\\ \cline{2-2} \cline{6-6} & \checkmark & & & & 35.93 \\\midrule 
\multirow{2}{*}{HFT-D} & -   & \multirow{2}{*}{LSS \cite{philion2020lift}} & \multirow{2}{*}{VPN \cite{pan2020cross}}  & \multirow{2}{*}{34.92M} & 32.51\\ \cline{2-2} \cline{6-6} & \checkmark & & & & 34.32\\\bottomrule
\end{tabular}
\tabcaption{Ablation study on the KITTI 3D Object dataset using different sub-modules. CBFT and CFFT indicate the category of the corresponding branch. MLS means mutual learning scheme.}
\label{tab:mutuallearning_table}
  \end{minipage}
 \hfill
\begin{minipage}[b]{0.41\linewidth}
\centering
\includegraphics[width=1.1\linewidth,trim= 0 5 0 25,clip]{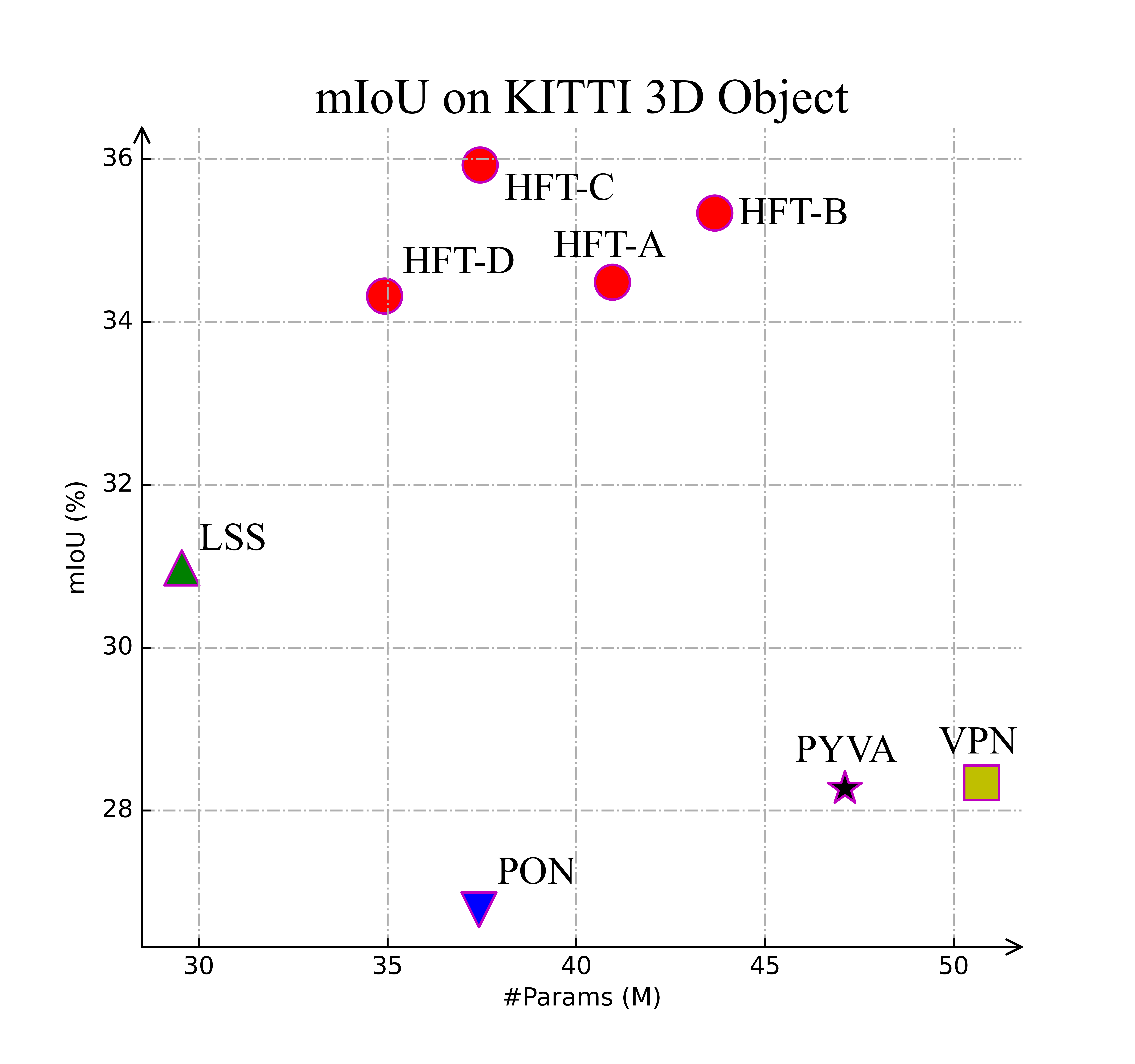}
\figcaption{The number of parameters and performance of different methods on the KITTI 3D Object test set.}
\label{fig:mutuallearning_fig}
\end{minipage}
 
\end{figtab}

\begin{figure*}[htp]
    \centering
    \setlength\tabcolsep{.05cm}
    \begin{tabular}{cc}
        \includegraphics[height=4.0cm]{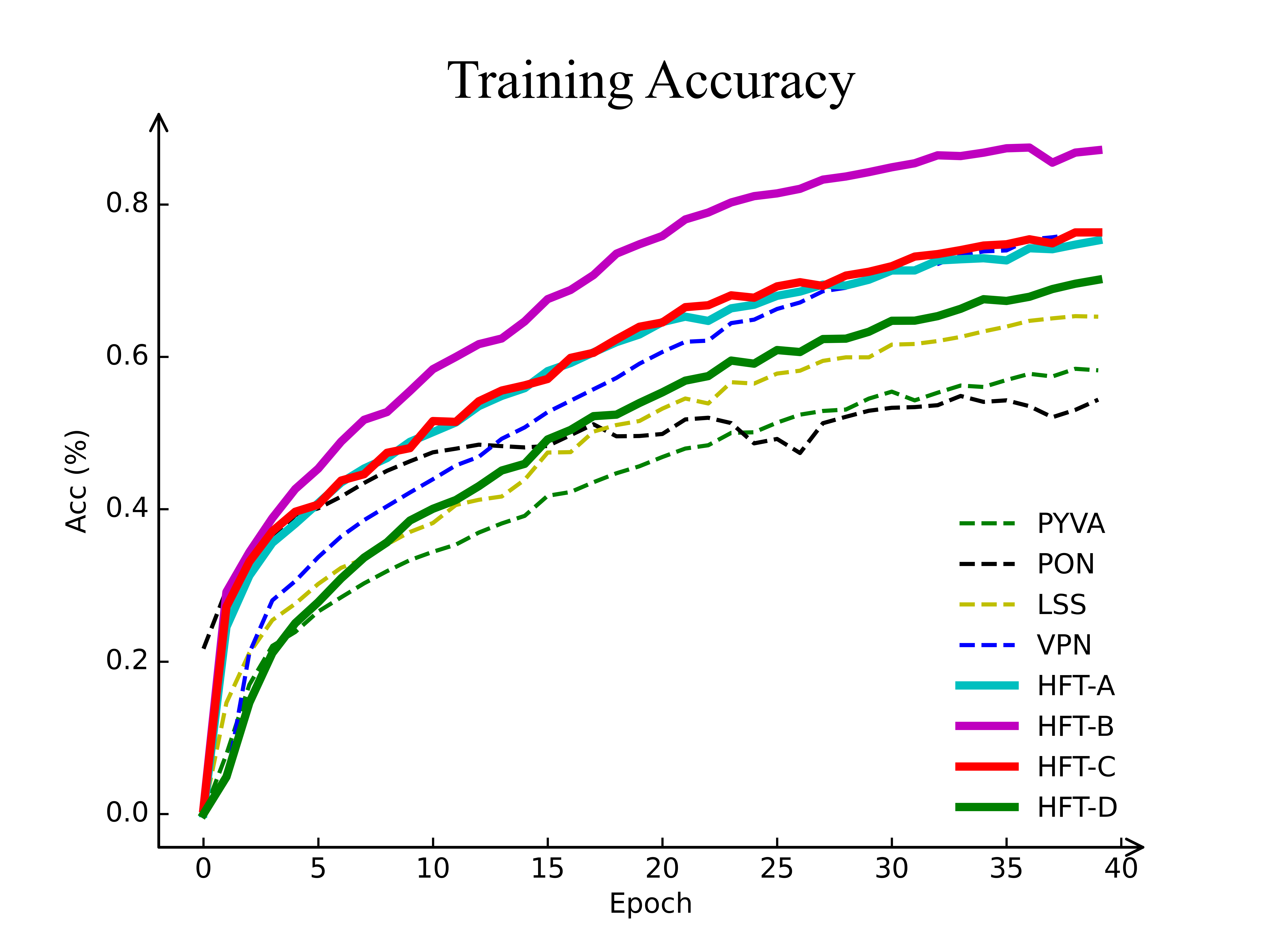}&
        \includegraphics[height=4.0cm]{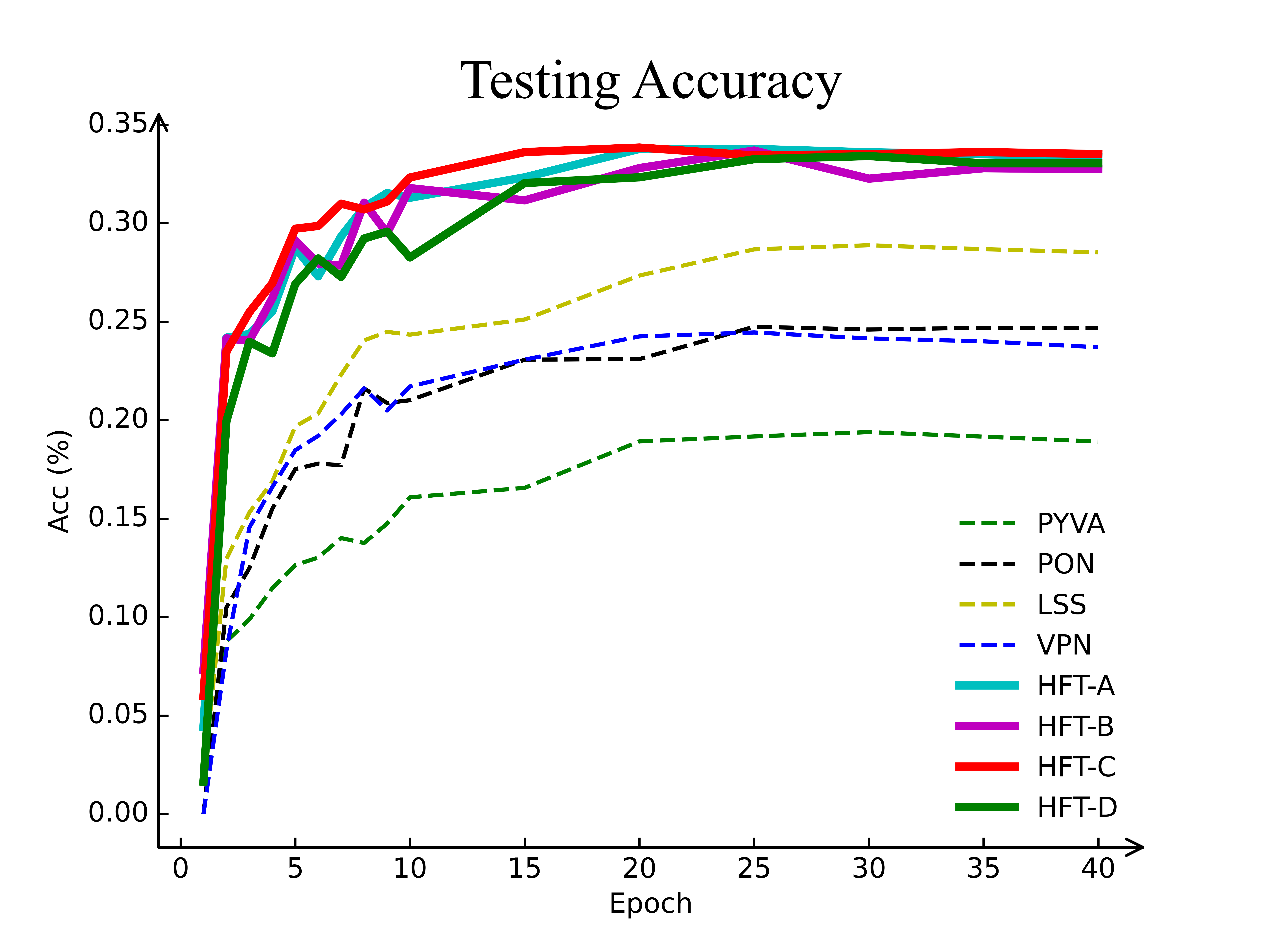}\\
        \includegraphics[height=4.0cm]{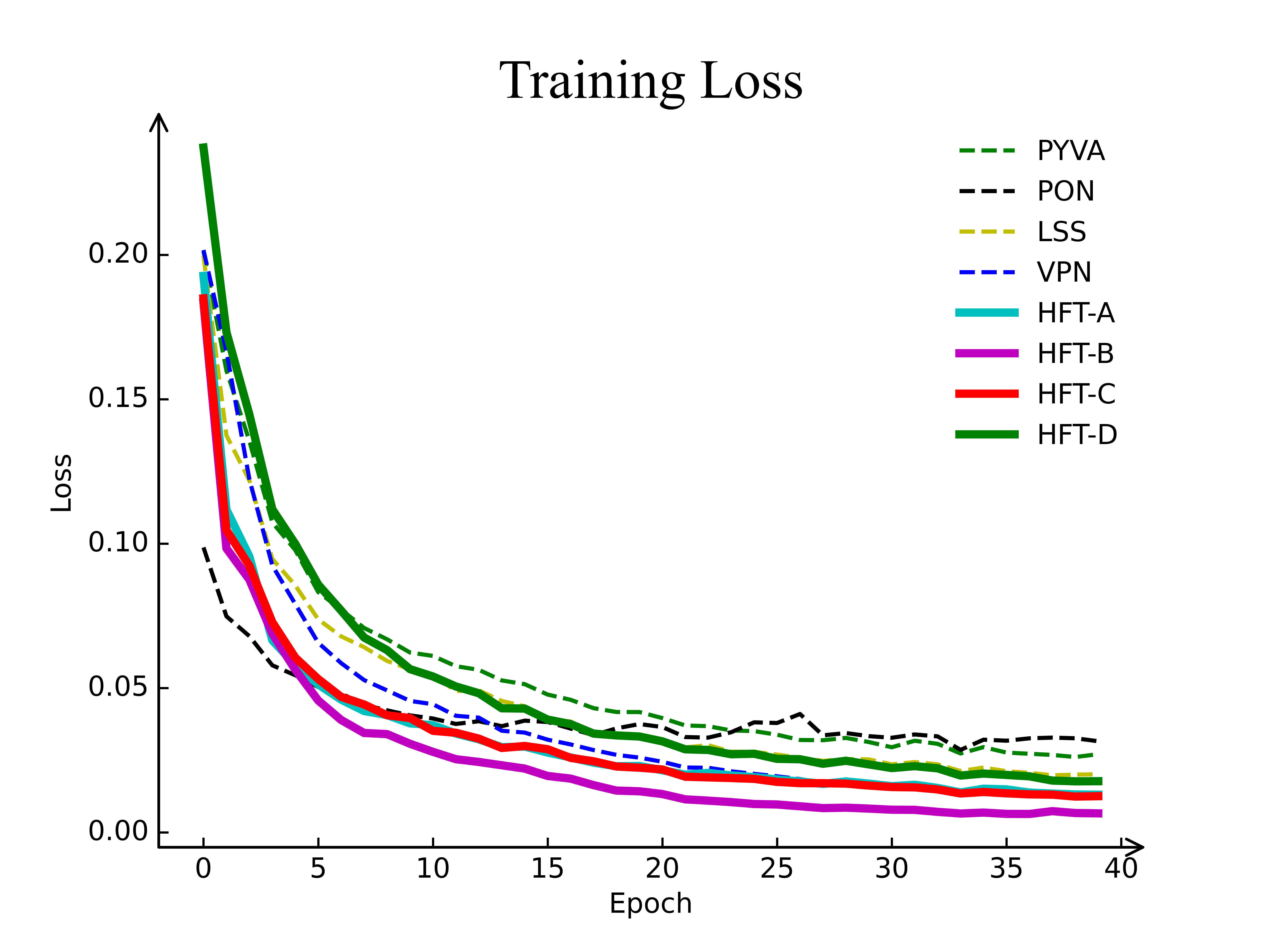}&
        \includegraphics[height=4.0cm]{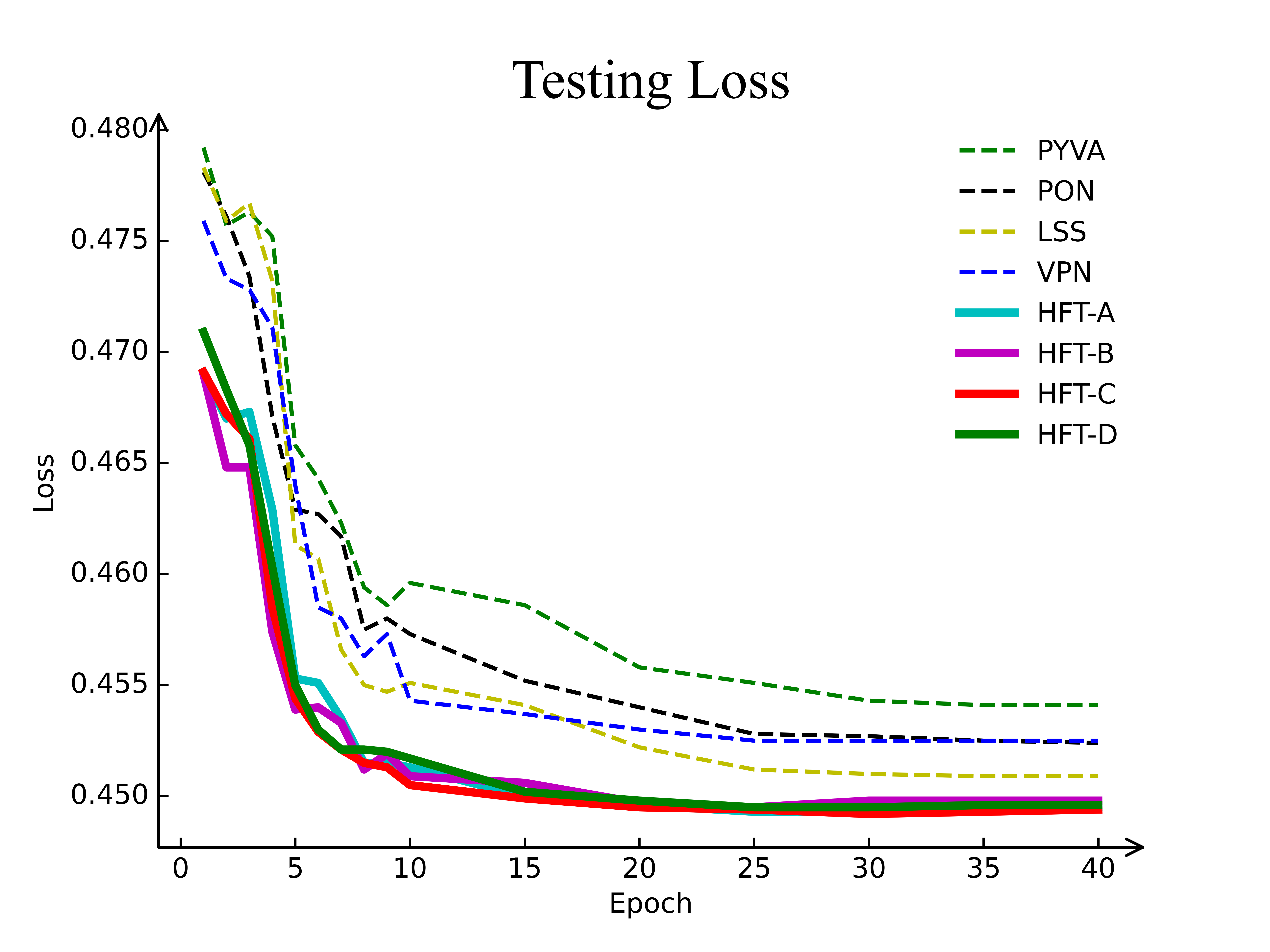}\\
    \end{tabular}
    \caption{The evolution of accuracy and loss in the training and testing process on the KITTI 3D Object set. The configurations are defined in \figref{mutuallearning_fig}.}
    \label{fig:accuracy_loss}
\end{figure*}

\begin{itemize}
    \item[$\bullet$] \textbf{Overall Performance}. As illustrated in \figref{mutuallearning_fig}, HFT networks with different configurations lead semantic segmentation performance among baselines. With negligible extra computing budget, HFT models achieve much higher mIoU scores. In other words, HFT keeps both high performance and high efficiency. Compared with VPN \cite{pan2020cross}, HFT-D (with PON \cite{roddick2020predicting}-style Geometric Transformer and VPN\cite{pan2020cross}-style Global Transformer) model reduces the number of parameters by 31.2\% from 50.74M to 34.92M, but it still obtains a relative 21.1\% improvement of mIoU. \par
    
    \item[$\bullet$] \textbf{Feature Fusion and Mutual Learning}. As listed in \tabref{mutuallearning_table}, HFT model without mutual learning scheme (MLS) simply fuses the BEV features generated by CBFT and CFFT. HFT obtains segmentation accuracy improvement by feature fusion. Employing MLS boosts the performance of HFT framework, which reveals the effectiveness of MLS.\par
    
    \item[$\bullet$] \textbf{Robustness and Generalization}. As illustrated in \figref{accuracy_loss}. HFT model surpasses both the corresponding CBFT and CFFT during the testing process. It's worth noting that VPN \cite{pan2020cross} achieves higher training accuracy and smaller training loss, but it fails to manifest better performance than HFT model during the testing stage. When dealing with novel autonomous driving scenarios, HFT keeps better robustness and generalization ability.\par
\end{itemize}

\textbf{Mutual Learning Schemes.} In order to exploit the effectiveness of different mutual learning, we construct four different configurations described as follows. We use PON \cite{roddick2020predicting} as the baseline method. Quantitative results are listed in \tabref{nuscenes ablation on kd}. 
\begin{table*}[]
\centering
\caption{Ablation study of mutual learning schemes on the nuScenes dataset.}
\label{tab:nuscenes ablation on kd}
\scriptsize
\begin{tabular}{@{}c|cccccccccccccc|c@{}}
\toprule
Method & \rot{Drivable} & \rot{Ped. crossing} & \rot{Walkway} & \rot{Carpark} & \rot{Car} & \rot{Truck} & \rot{Bus} & \rot{Trailer} & \rot{Constr. veh.} & \rot{Pedestrian} & \rot{Motorcycle} & \rot{Bicycle} & \rot{Traf. Cone} & \rot{Barrier} & \rot{Mean} \\
\midrule
baseline & \textbf{56.2} & 26.4 & 32.2 & 21.3 & 19.3 & 13.2 & 21.4 & 12.5 & 7.4 & 4.2 & 3.5 & 4.3 & 2.0 & 6.3 & 16.4 \\
\midrule
CBFT-teacher & 53.5 & 25.0 & 32.8 & 21.7 & 19.6 & 12.6 & 21.2 & 12.3 &  7.2 & 4.3 & 2.6 & 3.8 & 1.7 & 5.5 & 16.0 \\
CFFT-teacher & 53.6 & 27.7 & 32.8 & 23.0 & 25.5 & 13.8 & 21.7 & 12.3 & \textbf{7.5} & 5.4 & 3.5 & 3.9 & 2.9 & 8.1 & 17.3  \\
output sim.  & 53.3 & 27.8 & 33.6 & \textbf{24.4} & 25.0 & 15.7 & \textbf{24.3} & \textbf{15.3} & 7.3 & \textbf{5.5} & \textbf{5.0} & \textbf{6.9} & \textbf{3.8} & \textbf{9.9} & \textbf{18.4} \\
sub-feature sim. & \textbf{53.4} & \textbf{29.5} & \textbf{34.0} &  22.9 & 24.6 & \textbf{15.7} & 24.1 & 14.4 & 8.6 & 5.0 & 3.9 & 6.1 & 3.1 & 8.6 & 18.1 \\
\bottomrule
\end{tabular}
\end{table*}
\begin{itemize}
    \item[$\bullet$] Using CBFT model as the distillation teacher (CBFT-teacher) fails to achieve better results than baseline, with a 16.0\% score of mIoU.
    \item[$\bullet$] To use CFFT model as the distillation teacher (CFFT-teacher) outperforms baseline by 0.9\%, reaching a 17.3\% score of mIoU. We induce that compared with the lack of geometric priors in CFFT models, the flat-world assumption of CBFT models has a more negative impact on the semantic segmentation accuracy. 
    \item[$\bullet$] Setting regularization on the output (output sim.) of two modules offers a positive impact on small and rare classes, \textit{e}.\textit{g}., pedestrian and barrier. The performance of this mutual learning configuration achieves the best performance among the listed mutual learning configurations, which is 2.0\% higher than baseline under the mIoU metric.
    \item[$\bullet$] By setting regularization on sub-features (sub-feature sim.) which stand for the semantic maps in BEV of different depth ranges, HFT achieves the highest performance on drivable area, pedestrian crossing, walkway and truck among these four mutual learning configurations. 
\end{itemize}
\quad In summary, the ablation study verifies the effectiveness of mutual learning schemes in HFT framework.\par
\begin{figure*}[htp]
    \centering
    \setlength\tabcolsep{.05cm}
    \begin{tabular}{ccccc}
        \includegraphics[height=1.9cm]{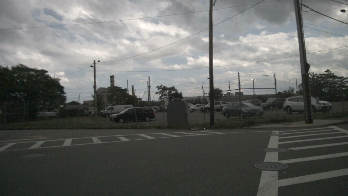}&
        \includegraphics[height=1.9cm]{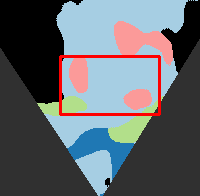}&
        \includegraphics[height=1.9cm]{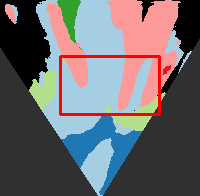}&
        \includegraphics[height=1.9cm]{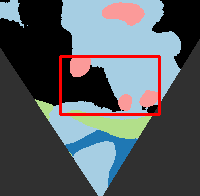}&
        \includegraphics[height=1.9cm]{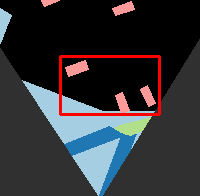}\\
        
        \includegraphics[height=1.9cm]{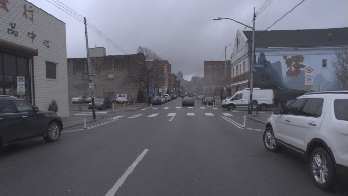}&
        \includegraphics[height=1.9cm]{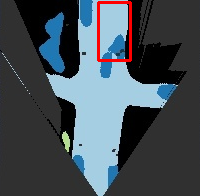}&
        \includegraphics[height=1.9cm]{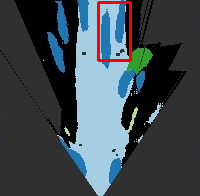}&
        \includegraphics[height=1.9cm]{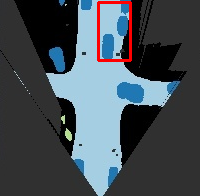}&
        \includegraphics[height=1.9cm]{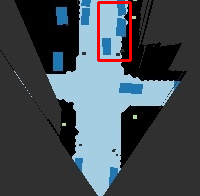}\\
        
        \includegraphics[height=1.9cm]{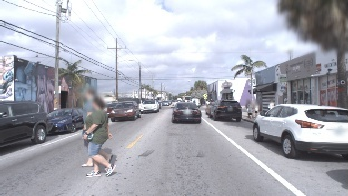}&
        \includegraphics[height=1.9cm]{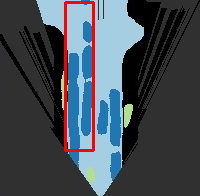}&
        \includegraphics[height=1.9cm]{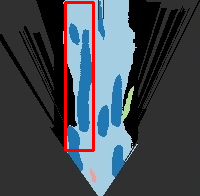}&
        \includegraphics[height=1.9cm]{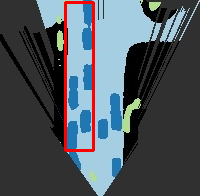}&
        \includegraphics[height=1.9cm]{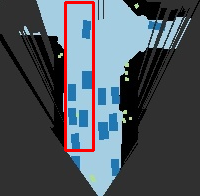}\\
        FV Image & PYVA \cite{yang2021projecting} & PON \cite{roddick2020predicting} & HFT & Ground Truth \\ 
    \end{tabular}
    \caption{Qualitative results on the nuScenes (the 1st row) and Argoverse (the 2nd and 3rd row) dataset. For each grid location $i$, we visualize the class with the largest index $c$ which has occupancy probability $p_i > 0.5$. Black regions (outside field of view or no lidar returns) are ignored during evaluation.}
    \label{fig:nuscenes}
\end{figure*}

\subsection{Qualitative Results}
\quad In this part, we illustrate some results generated by different methods. We choose three FV images in these datasets and list the FV image, semantic segmentation map in BEV produced by PYVA \cite{yang2021projecting}, PON \cite{roddick2020predicting} and HFT, ground truth semantic annotations from left to right respectively.\par

As listed in \figref{nuscenes} , HFT can produce more accurate semantic segmentation results with a large improvement in visualization than PYVA \cite{yang2021projecting} and PON \cite{roddick2020predicting}. It's worth noting that HFT can predict the layout of roads, vehicles and other kinds of objects especially in uphill, downhill and uneven scenes, which is usually the hard cases for the existing CBFT and CFFT. Another prominent promotion is that HFT can produce much more clear edges in dense scenes.\par

\section{Conclusion}
\quad In this paper, we have proposed a novel framework with HFT to lift perspective representations from monocular images. We explore the vital differences between CBFT and CFFT, concluding that geometric priors from the former and global spatial correlation from the latter are both vital for semantic perception in BEV. Inspired by both influential factors, we propose the HFT framework comprising both categories of methods and employ a mutual learning scheme to push both branches to learn reciprocally from each other. Extensive experiments on five challenging benchmarks demonstrate that HFT substantially outperforms the previous state-of-the-art method by relatively 13.3\% on the Argoverse dataset and 16.8\% on the KITTI 3D Object datasets. With negligible computing budget, HFT keeps both high segmentation performance and high computing efficiency, which is meaningful for visual perception in autonomous driving.\par

\clearpage
\appendix
\section{Additional Ablation Study}
\quad \textbf{Additional Ablation Study on Loss Function} To exploit the effects of different loss functions in mutual learning, we select L1 Loss, KL Divergence Loss and L2 Loss for comparison. As illustrated in \tabref{nuscenes ablation on loss}, the ablation experimental results show that the selection of proper loss function between Geometric Transformer and Global Transformer has an impact on the final performance of HFT model. Compared with L1 Loss and KL Loss, the HFT model with L2 Loss function ranks first on 12 over 14 classes on nuScenes \cite{caesar2020nuscenes} dataset, \textit{i}.\textit{e}. 21.3\% score of mIoU, which is a large 10.4\% relative improvement than baseline (19.3\% score of mIoU).\par

\section{More Qualitative Results}
\quad In this part, we illustrate more results generated by different methods. The listed images are sampled from nuScenes \cite{caesar2020nuscenes}, Argoverse \cite{chang2019argoverse}, KITTI Raw, KITTI Odometry, and KITTI 3D Object \cite{geiger2012we,behley2019semantickitti} datasets. We choose several FV images in these datasets and list the FV image, semantic segmentation map in BEV produced by PYVA \cite{yang2021projecting}, PON \cite{roddick2020predicting}, HFT, and ground truth semantic annotations from left to right respectively.\par
\begin{figure*}[htp]
    \centering
    \setlength\tabcolsep{.05cm}
    \begin{tabular}{ccccc}
        \includegraphics[height=1.9cm]{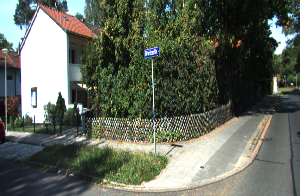}&
        \includegraphics[height=1.9cm]{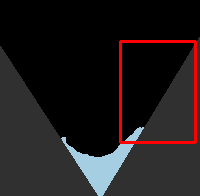}&
        \includegraphics[height=1.9cm]{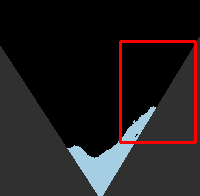}&
        \includegraphics[height=1.9cm]{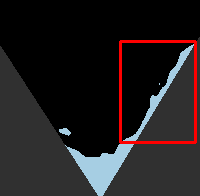}&
        \includegraphics[height=1.9cm]{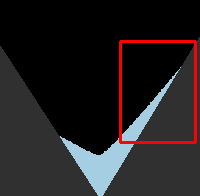}\\
        
        \includegraphics[height=1.9cm]{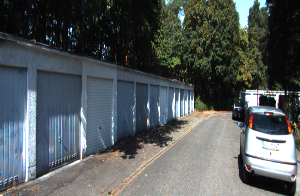}&
        \includegraphics[height=1.9cm]{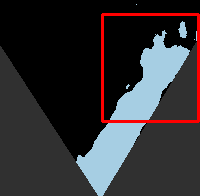}&
        \includegraphics[height=1.9cm]{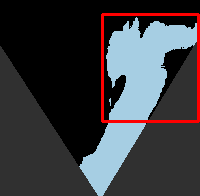}&
        \includegraphics[height=1.9cm]{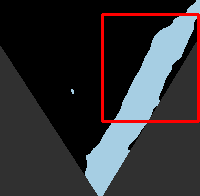}&
        \includegraphics[height=1.9cm]{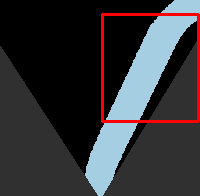}\\
    
        \includegraphics[height=1.9cm]{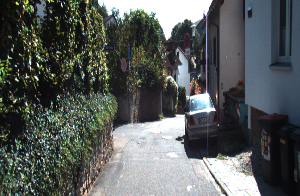}&
        \includegraphics[height=1.9cm]{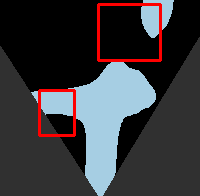}&
        \includegraphics[height=1.9cm]{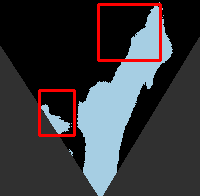}&
        \includegraphics[height=1.9cm]{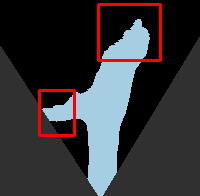}&
        \includegraphics[height=1.9cm]{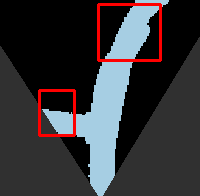}\\
        
        \includegraphics[height=1.9cm]{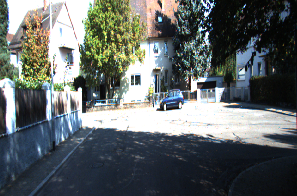}&
        \includegraphics[height=1.9cm]{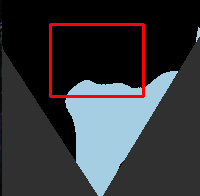}&
        \includegraphics[height=1.9cm]{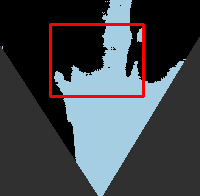}&
        \includegraphics[height=1.9cm]{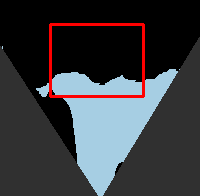}&
        \includegraphics[height=1.9cm]{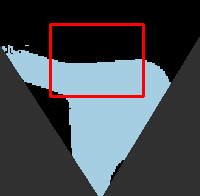}\\
        
        \includegraphics[height=1.9cm]{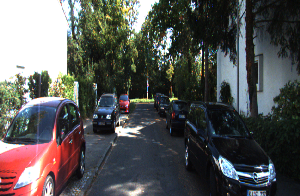}&
        \includegraphics[height=1.9cm]{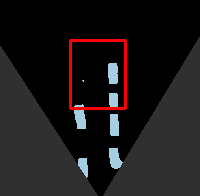}&
        \includegraphics[height=1.9cm]{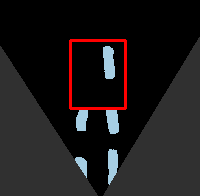}&
        \includegraphics[height=1.9cm]{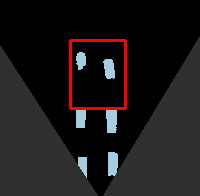}&
        \includegraphics[height=1.9cm]{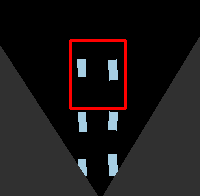}\\
        
        \includegraphics[height=1.9cm]{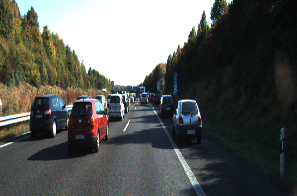}&
        \includegraphics[height=1.9cm]{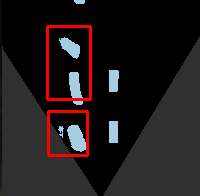}&
        \includegraphics[height=1.9cm]{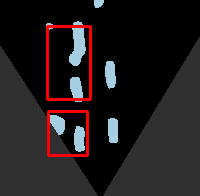}&
        \includegraphics[height=1.9cm]{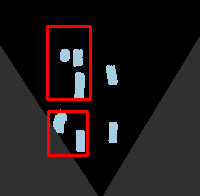}&
        \includegraphics[height=1.9cm]{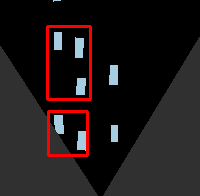}\\
        FV Image & PYVA \cite{yang2021projecting} & PON \cite{roddick2020predicting} & HFT & Ground Truth \\ 
    \end{tabular}
    \caption{Qualitative results on the KITTI Raw (the 1st and 2nd row), KITTI Odometry (the 3rd and 4th row), and KITTI 3D Object (the 5th and 6th row).}
    \label{fig:Kittirawodometryobject}
\end{figure*}

As listed in \figref{nuscenes} and \figref{Kittirawodometryobject}, the improvement can be summarized as follows.\par 
\begin{itemize}
    \item[$\bullet$] HFT produces more accurate semantic segmentation results with a large improvement in visualization than PYVA \cite{yang2021projecting} and PON \cite{roddick2020predicting}.
    \item[$\bullet$] HFT predicts the layout of roads, vehicles, and other kinds of dynamic objects, especially in uphill, downhill, and uneven scenes. BEV  semantic segmentation in these scenes is usually the hard case for CBFT and CFFT.
    \item[$\bullet$] HFT produces much more clear edges in dense scenes, which is meaningful in crowded autonomous driving circumstances.
\end{itemize}
\quad For each grid location $i$, we visualize the class with the largest index $c$ which has occupancy probability $p_i > 0.5$. Black regions (outside field of view or no lidar returns) are ignored during evaluation.
\begin{table*}[t!]
\centering
\caption{Ablation study of loss function on the nuScenes dataset.}
\label{tab:nuscenes ablation on loss}
\begin{tabular}{@{}c|cccccccccccccc|c@{}}
\toprule
Method & \rot{Drivable} & \rot{Ped. crossing} & \rot{Walkway} & \rot{Carpark} & \rot{Car} & \rot{Truck} & \rot{Bus} & \rot{Trailer} & \rot{Constr. veh.} & \rot{Pedestrian} & \rot{Motorcycle} & \rot{Bicycle} & \rot{Traf. Cone} & \rot{Barrier} & \rot{Mean} \\
\midrule
L1 Loss & 53.1 & 31.4 & 32.4 & 21.0 & 26.4 & 16.5 & 24.8 & 15.9 & \textbf{5.8} &  6.0 & 8.5 & 8.8 & \textbf{5.5} & \textbf{11.8} & 19.1 \\
KL Loss  & 54.8 & 34.3 & 33.7 & 21.1 & 29.8 & 17.8 & 27.0 & 16.5 & 4.9 & 7.7 & 10.6 & 11.0 & 5.4 & 13.4 & 20.6 \\
L2 Loss & \textbf{55.9} & \textbf{35.6} & \textbf{35.4} & \textbf{23.2} & \textbf{30.6} & \textbf{19.1} & \textbf{27.3} & \textbf{18.6} & \textbf{5.8} & \textbf{7.9} & \textbf{11.07} & \textbf{11.7} & 5.1 & 11.4 & \textbf{21.3} \\
\bottomrule
\end{tabular}
\end{table*}
\begin{figure*}[htp]
    \centering
    \setlength\tabcolsep{.05cm}
    \begin{tabular}{ccccc}
        
        \includegraphics[height=1.9cm]{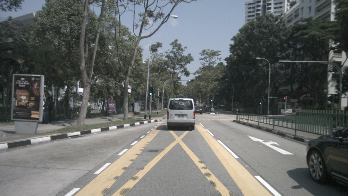}&
        \includegraphics[height=1.9cm]{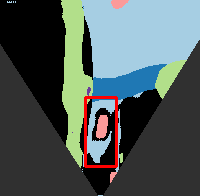}&
        \includegraphics[height=1.9cm]{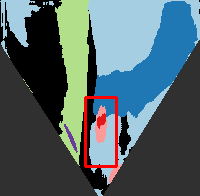}&
        \includegraphics[height=1.9cm]{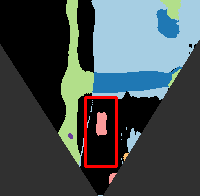}&
        \includegraphics[height=1.9cm]{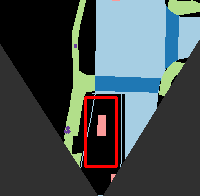}\\
        
        \includegraphics[height=1.9cm]{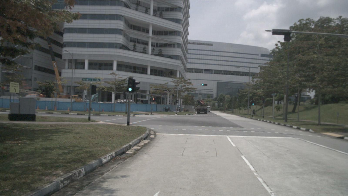}&
        \includegraphics[height=1.9cm]{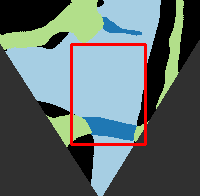}&
        \includegraphics[height=1.9cm]{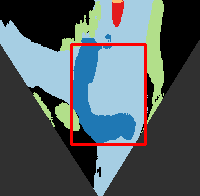}&
        \includegraphics[height=1.9cm]{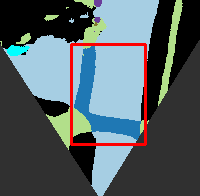}&
        \includegraphics[height=1.9cm]{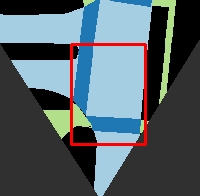}\\
        
        \includegraphics[height=1.9cm]{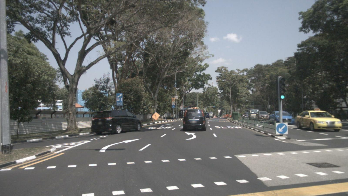}&
        \includegraphics[height=1.9cm]{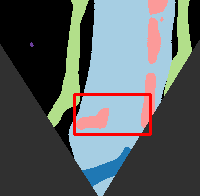}&
        \includegraphics[height=1.9cm]{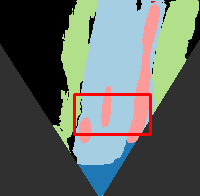}&
        \includegraphics[height=1.9cm]{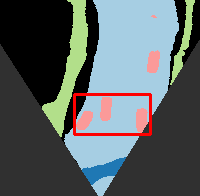}&
        \includegraphics[height=1.9cm]{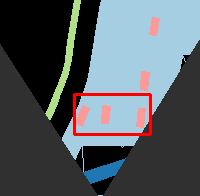}\\
        
        \includegraphics[height=1.9cm]{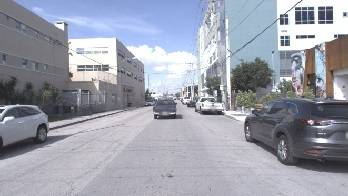}&
        \includegraphics[height=1.9cm]{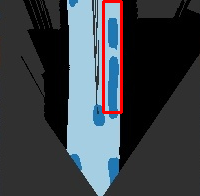}&
        \includegraphics[height=1.9cm]{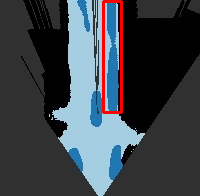}&
        \includegraphics[height=1.9cm]{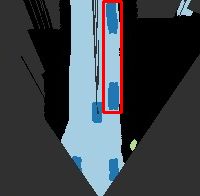}&
        \includegraphics[height=1.9cm]{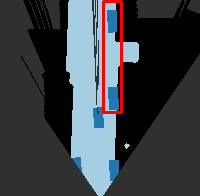}\\
        
        \includegraphics[height=1.9cm]{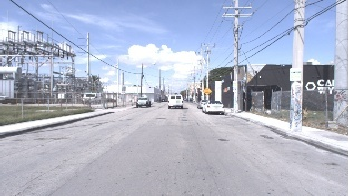}&
        \includegraphics[height=1.9cm]{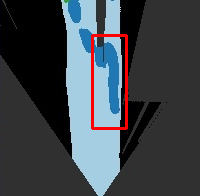}&
        \includegraphics[height=1.9cm]{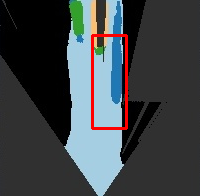}&
        \includegraphics[height=1.9cm]{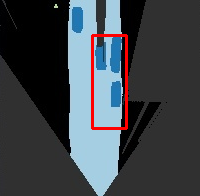}&
        \includegraphics[height=1.9cm]{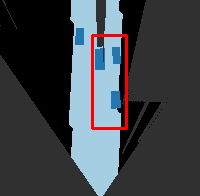}\\
        
        \includegraphics[height=1.9cm]{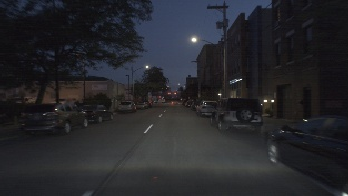}&
        \includegraphics[height=1.9cm]{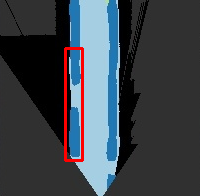}&
        \includegraphics[height=1.9cm]{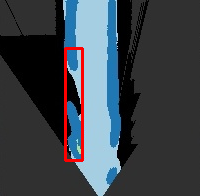}&
        \includegraphics[height=1.9cm]{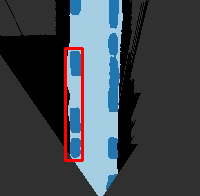}&
        \includegraphics[height=1.9cm]{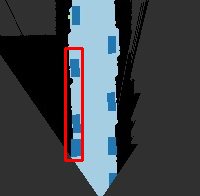}\\

        FV Image & PYVA \cite{yang2021projecting} & PON \cite{roddick2020predicting} & HFT & Ground Truth \\ 
    \end{tabular}
    \caption{Qualitative results on the nuScenes (the 1st, 2nd, and 3rd row) and Argoverse (the 4th, 5th, and 6th row) dataset.}
    \label{fig:nuscenes}
\end{figure*}

\clearpage
%
%

\bibliographystyle{splncs04}
\bibliography{egbib}

\begin{thebibliography}{10}
\providecommand{\url}[1]{\texttt{#1}}
\providecommand{\urlprefix}{URL }
\providecommand{\doi}[1]{https://doi.org/#1}

\bibitem{coo}
Batra, T., Parikh, D.: Cooperative learning with visual attributes. arXiv
  preprint arXiv:1705.05512  (2017)

\bibitem{behley2019semantickitti}
Behley, J., Garbade, M., Milioto, A., Quenzel, J., Behnke, S., Stachniss, C.,
  Gall, J.: Semantickitti: A dataset for semantic scene understanding of lidar
  sequences. In: IEEE International Conference on Computer Vision (2019)

\bibitem{beltran2018birdnet}
Beltr{\'a}n, J., Guindel, C., Moreno, F.M., Cruzado, D., Garcia, F.,
  De~La~Escalera, A.: Birdnet: a 3d object detection framework from lidar
  information. In: International Conference on Intelligent Transportation
  Systems (2018)

\bibitem{caesar2020nuscenes}
Caesar, H., Bankiti, V., Lang, A.H., Vora, S., Liong, V.E., Xu, Q., Krishnan,
  A., Pan, Y., Baldan, G., Beijbom, O.: nuscenes: A multimodal dataset for
  autonomous driving. In: IEEE Conference on Computer Vision and Pattern
  Recognition (2020)

\bibitem{chabot2017deep}
Chabot, F., Chaouch, M., Rabarisoa, J., Teuliere, C., Chateau, T.: Deep manta:
  A coarse-to-fine many-task network for joint 2d and 3d vehicle analysis from
  monocular image. In: IEEE Conference on Computer Vision and Pattern
  Recognition (2017)

\bibitem{chang2019argoverse}
Chang, M.F., Lambert, J., Sangkloy, P., Singh, J., Bak, S., Hartnett, A., Wang,
  D., Carr, P., Lucey, S., Ramanan, D., et~al.: Argoverse: 3d tracking and
  forecasting with rich maps. In: IEEE Conference on Computer Vision and
  Pattern Recognition (2019)

\bibitem{chen2016monocular}
Chen, X., Kundu, K., Zhang, Z., Ma, H., Fidler, S., Urtasun, R.: Monocular 3d
  object detection for autonomous driving. In: IEEE Conference on Computer
  Vision and Pattern Recognition (2016)

\bibitem{chen2017multi}
Chen, X., Ma, H., Wan, J., Li, B., Xia, T.: Multi-view 3d object detection
  network for autonomous driving. In: IEEE conference on Computer Vision and
  Pattern Recognition (2017)

\bibitem{mmseg2020}
Contributors, M.: {MMSegmentation}: Openmmlab semantic segmentation toolbox and
  benchmark. \url{https://github.com/open-mmlab/mmsegmentation} (2020)

\bibitem{3D1}
Ding, M., Huo, Y., Yi, H., Wang, Z., Shi, J., Lu, Z., Luo, P.: Learning
  depth-guided convolutions for monocular 3d object detection. IEEE Conference
  on Computer Vision and Pattern Recognition Workshops  (2020)

\bibitem{du2018general}
Du, X., Ang, M.H., Karaman, S., Rus, D.: A general pipeline for 3d detection of
  vehicles. In: IEEE International Conference on Robotics and Automation (2018)

\bibitem{multilayer}
Gardner, M.W., Dorling, S.: Artificial neural networks (the multilayer
  perceptron)—a review of applications in the atmospheric sciences.
  Atmospheric environment  (1998)

\bibitem{dev1}
Geiger, A., Lauer, M., Wojek, C., Stiller, C., Urtasun, R.: 3d traffic scene
  understanding from movable platforms. IEEE Transactions on Pattern Analysis
  and Machine Intelligence  (2014)

\bibitem{geiger2012we}
Geiger, A., Lenz, P., Urtasun, R.: Are we ready for autonomous driving? the
  kitti vision benchmark suite. In: IEEE Conference on Computer Vision and
  Pattern Recognition (2012)

\bibitem{dev2}
Gupta, S., Tolani, V., Davidson, J., Levine, S., Sukthankar, R., Malik, J.:
  Cognitive mapping and planning for visual navigation. International Journal
  of Computer Vision  (2019)

\bibitem{dual}
He, D., Xia, Y., Qin, T., Wang, L., Yu, N., Liu, T.Y., Ma, W.Y.: Dual learning
  for machine translation. Advances in neural information processing systems
  (2016)

\bibitem{hinton2015distilling}
Hinton, G., Vinyals, O., Dean, J.: Distilling the knowledge in a neural
  network. arXiv preprint arXiv:1503.02531  (2015)

\bibitem{BC0}
Hong, J., Sapp, B., Philbin, J.: Rules of the road: Predicting driving behavior
  with a convolutional model of semantic interactions. IEEE Conference on
  Computer Vision and Pattern Recognition  (2019)

\bibitem{hu2021fiery}
Hu, A., Murez, Z., Mohan, N., Dudas, S., Hawke, J., Badrinarayanan, V.,
  Cipolla, R., Kendall, A.: Fiery: Future instance prediction in bird's-eye
  view from surround monocular cameras. In: IEEE International Conference on
  Computer Vision (2021)

\bibitem{kim2021feature}
Kim, J., Hyun, M., Chung, I., Kwak, N.: Feature fusion for online mutual
  knowledge distillation. In: International Conference on Pattern Recognition
  (2021)

\bibitem{BC1}
Kim, J., Moon, S., Rohrbach, A., Darrell, T., Canny, J.F.: Advisable learning
  for self-driving vehicles by internalizing observation-to-action rules. IEEE
  Conference on Computer Vision and Pattern Recognition  (2020)

\bibitem{lan2018knowledge}
Lan, X., Zhu, X., Gong, S.: Knowledge distillation by on-the-fly native
  ensemble. arXiv preprint arXiv:1806.04606  (2018)

\bibitem{li2016vehicle}
Li, B., Zhang, T., Xia, T.: Vehicle detection from 3d lidar using fully
  convolutional network. arXiv preprint arXiv:1608.07916  (2016)

\bibitem{3D2}
Li, B., Ouyang, W., Sheng, L., Zeng, X., Wang, X.: Gs3d: An efficient 3d object
  detection framework for autonomous driving. IEEE Conference on Computer
  Vision and Pattern Recognition  (2019)

\bibitem{li2021hdmapnet}
Li, Q., Wang, Y., Wang, Y., Zhao, H.: Hdmapnet: An online hd map construction
  and evaluation framework. arXiv e-prints  (2021)

\bibitem{focal}
Lin, T.Y., Goyal, P., Girshick, R., He, K., Doll{\'a}r, P.: Focal loss for
  dense object detection. In: Proceedings of the IEEE international conference
  on computer vision (2017)

\bibitem{liu2021swin}
Liu, Z., Lin, Y., Cao, Y., Hu, H., Wei, Y., Zhang, Z., Lin, S., Guo, B.: Swin
  transformer: Hierarchical vision transformer using shifted windows. arXiv
  preprint arXiv:2103.14030  (2021)

\bibitem{loshchilov2018fixing}
Loshchilov, I., Hutter, F.: Fixing weight decay regularization in adam  (2018)

\bibitem{lu2019monocular}
Lu, C., van~de Molengraft, M.J.G., Dubbelman, G.: Monocular semantic occupancy
  grid mapping with convolutional variational encoder--decoder networks. IEEE
  Robotics and Automation Letters  (2019)

\bibitem{BC2}
Ma, Y., Zhu, X., Zhang, S., Yang, R., Wang, W., Manocha, D.: Trafficpredict:
  Trajectory prediction for heterogeneous traffic-agents. In: AAAI Conference
  on Artificial Intelligence (2019)

\bibitem{dev4}
M{\'a}ttyus, G., Wang, S., Fidler, S., Urtasun, R.: Hd maps: Fine-grained road
  segmentation by parsing ground and aerial images. IEEE Conference on Computer
  Vision and Pattern Recognition  (2016)

\bibitem{dev5}
Mousavian, A., Anguelov, D., Flynn, J., Kosecka, J.: 3d bounding box estimation
  using deep learning and geometry. IEEE Conference on Computer Vision and
  Pattern Recognition  (2017)

\bibitem{BC3}
Mozaffari, S., Al-Jarrah, O.Y., Dianati, M., Jennings, P.A., Mouzakitis, A.:
  Deep learning-based vehicle behavior prediction for autonomous driving
  applications: A review. IEEE Transactions on Intelligent Transportation
  Systems  (2022)

\bibitem{pan2020cross}
Pan, B., Sun, J., Leung, H.Y.T., Andonian, A., Zhou, B.: Cross-view semantic
  segmentation for sensing surroundings. IEEE Robotics and Automation Letters
  (2020)

\bibitem{Paszke2019PyTorchAI}
Paszke, A., Gross, S., Massa, F., Lerer, A., Bradbury, J., Chanan, G., Killeen,
  T., Lin, Z., Gimelshein, N., Antiga, L., et~al.: Pytorch: An imperative
  style, high-performance deep learning library. Advances in neural information
  processing systems  (2019)

\bibitem{philion2020lift}
Philion, J., Fidler, S.: Lift, splat, shoot: Encoding images from arbitrary
  camera rigs by implicitly unprojecting to 3d. In: European Conference on
  Computer Vision (2020)

\bibitem{phuong2019distillation}
Phuong, M., Lampert, C.H.: Distillation-based training for multi-exit
  architectures. In: IEEE International Conference on Computer Vision (2019)

\bibitem{poirson2016fast}
Poirson, P., Ammirato, P., Fu, C.Y., Liu, W., Kosecka, J., Berg, A.C.: Fast
  single shot detection and pose estimation. In: International Conference on 3D
  Vision (2016)

\bibitem{qi2018frustum}
Qi, C.R., Liu, W., Wu, C., Su, H., Guibas, L.J.: Frustum pointnets for 3d
  object detection from rgb-d data. In: IEEE Conference on Computer Vision and
  Pattern Recognition (2018)

\bibitem{reiher2020sim2real}
Reiher, L., Lampe, B., Eckstein, L.: A sim2real deep learning approach for the
  transformation of images from multiple vehicle-mounted cameras to a
  semantically segmented image in bird’s eye view. In: International
  Conference on Intelligent Transportation Systems (2020)

\bibitem{roddick2020predicting}
Roddick, T., Cipolla, R.: Predicting semantic map representations from images
  using pyramid occupancy networks. In: IEEE Conference on Computer Vision and
  Pattern Recognition (2020)

\bibitem{3D3}
Roddick, T., Kendall, A., Cipolla, R.: Orthographic feature transform for
  monocular 3d object detection. In: British Machine Vision Conference (2019)

\bibitem{Russakovsky2015ImageNetLS}
Russakovsky, O., Deng, J., Su, H., Krause, J., Satheesh, S., Ma, S., Huang, Z.,
  Karpathy, A., Khosla, A., Bernstein, M.S., Berg, A.C., Fei-Fei, L.: Imagenet
  large scale visual recognition challenge. International Journal of Computer
  Vision  (2015)

\bibitem{schulter2018learning}
Schulter, S., Zhai, M., Jacobs, N., Chandraker, M.: Learning to look around
  objects for top-view representations of outdoor scenes. In: European
  Conference on Computer Vision (2018)

\bibitem{dev7}
Sengupta, S., Sturgess, P., Ladicky, L., Torr, P.H.S.: Automatic dense visual
  semantic mapping from street-level imagery. IEEE International Conference on
  Intelligent Robots and Systems  (2012)

\bibitem{3D4}
Simonelli, A., Bul{\`o}, S.R., Porzi, L., L{\'o}pez-Antequera, M.,
  Kontschieder, P.: Disentangling monocular 3d object detection. IEEE
  International Conference on Computer Vision  (2019)

\bibitem{attentionisall}
Vaswani, A., Shazeer, N., Parmar, N., Uszkoreit, J., Jones, L., Gomez, A.N.,
  Kaiser, {\L}., Polosukhin, I.: Attention is all you need. Advances in neural
  information processing systems  (2017)

\bibitem{dev8}
Wang, Z., Liu, B., Schulter, S., Chandraker, M.: A parametric top-view
  representation of complex road scenes. IEEE Conference on Computer Vision and
  Pattern Recognition  (2019)

\bibitem{wirges2018object}
Wirges, S., Fischer, T., Stiller, C., Frias, J.B.: Object detection and
  classification in occupancy grid maps using deep convolutional networks. In:
  International Conference on Intelligent Transportation Systems (2018)

\bibitem{yang2021projecting}
Yang, W., Li, Q., Liu, W., Yu, Y., Ma, Y., He, S., Pan, J.: Projecting your
  view attentively: Monocular road scene layout estimation via cross-view
  transformation. In: IEEE Conference on Computer Vision and Pattern
  Recognition (2021)

\bibitem{dev11}
Zhai, M., Bessinger, Z., Workman, S., Jacobs, N.: Predicting ground-level scene
  layout from aerial imagery. IEEE Conference on Computer Vision and Pattern
  Recognition  (2017)

\bibitem{Zhang2018DeepML}
Zhang, Y., Xiang, T., Hospedales, T.M., Lu, H.: Deep mutual learning. IEEE
  Conference on Computer Vision and Pattern Recognition  (2018)

\bibitem{dev9}
Zou, Q., Jiang, H., Dai, Q., Yue, Y., Chen, L., Wang, Q.: Robust lane detection
  from continuous driving scenes using deep neural networks. IEEE Transactions
  on Vehicular Technology  (2020)

\end{thebibliography}
\end{document}